\documentclass{article}
\usepackage[utf8]{inputenc}
\usepackage[authoryear, round]{natbib}

\title{Bayesian Causal Inference with Gaussian Process Networks}

\author{Enrico Giudice\footnote{Department of Mathematics and Computer Science, University of Basel, Basel, Switzerland} \and Jack Kuipers\footnote{Department of Biosystems Science and Engineering, ETH Zurich, Basel, Switzerland} \and Giusi Moffa$^{*,\,}$\footnote{Division of Psychiatry, University College London, London, UK}}
\date{\today}

\usepackage{amsmath, amsfonts}
\usepackage{graphicx}
\usepackage[a4paper, total={15cm, 22.2cm}]{geometry}
\usepackage{url}
\usepackage{algorithm}      
\usepackage[noend]{algpseudocode}
\usepackage{caption, subcaption}
\usepackage{hyperref}
\usepackage{mathtools}
\usepackage{bm}

\algnewcommand{\LineComment}[1]{ \hfill  \eqparbox{Comment}{$\triangleright$ #1}}
\newcommand{\given}{\,\vert\,}
\newcommand*{\boldone}{\text{\usefont{U}{bbold}{m}{n}1}}
\DeclareMathOperator*{\argmax}{arg\,max}

\begin{document}

\maketitle

\begin{abstract}
    Causal discovery and inference from observational data is an essential problem in statistics posing both modeling and computational challenges. These are typically addressed by imposing strict assumptions on the joint distribution such as linearity. We consider the problem of the Bayesian estimation of the effects of hypothetical interventions in the Gaussian Process Network (GPN) model, a flexible causal framework which allows describing the causal relationships nonparametrically. We detail how to perform causal inference on GPNs by simulating the effect of an intervention across the whole network and propagating the effect of the intervention on downstream variables. We further derive a simpler computational approximation by estimating the intervention distribution as a function of local variables only, modeling the conditional distributions via additive Gaussian processes. We extend both frameworks beyond the case of a known causal graph, incorporating uncertainty about the causal structure via Markov chain Monte Carlo methods. Simulation studies show that our approach is able to identify the effects of hypothetical interventions with non-Gaussian, non-linear observational data and accurately reflect the posterior uncertainty of the causal estimates. Finally we compare the results of our GPN-based causal inference approach to existing methods on a dataset of \textit{A.\ thaliana} gene expressions.
\end{abstract}
\vspace{0.5cm}

\section{Introduction}
Quantifying the causal relationships from purely observational data between variables in a system is a problem that has attracted great attention in the fields of statistics and machine learning. Full knowledge of the causal relations allows predicting the outcome of direct manipulations on the system, which can generally only be known from interventional data obtained by performing experiments such as randomized controlled trials \citep{Eberhardt2007}. 
Predicting the effect of such manipulations without the need of costly or infeasible experiments is of great practical relevance, specifically in the fields of computational biology \citep{Sachs2005}, medicine \citep{Richens2020} or AI \citep{Schollcausality}, since a central question concerns how a complex system will react to some treatment or outside influence of the user. 

Pearl's rules of $\textrm{do}$-calculus \citep{Pearl2000} allow computing the \textit{intervention} distributions resulting from these external manipulations from the joint distribution of the set of random variables together with a Directed Acyclic Graph (DAG). The DAG represents the qualitative causal relationships among the variables; each node in the graph represents a variable and a directed edge indicates a direct causal effect. Probabilistic models that are based on such DAGs, commonly called causal Bayesian Networks (BNs), provide conventional grounds for probabilistic causal inference, due to their compact representation of the joint distribution and their intuitive graphical description of the causal structure.

When the DAG is known, inference on causal BNs boils down to inference on the parameters of the joint distribution. In high-dimensional cases or general situations of uncertainty regarding the pattern of causal relations, the causal graph is however unknown and must therefore be estimated from the data. The DAG can be identified via targeted interventions \citep{cooper1999, GPNinterventions}, while for observational data, the problem of estimating the DAG is generally computationally difficult \citep{chickering2004large}, also due to the super-exponential growth of the number of DAGs with the number of nodes. 

Methods for causal inference from purely observational data typically employ structure learning algorithms to first make inference on the causal graph and then estimate intervention effects conditionally on the estimated DAG. Examples of such structure learning algorithms include constraint-based methods, such as the PC algorithm \citep{pc}, or Bayesian structure learning algorithms such as order \citep{BeingBayesian} or partition MCMC \citep{partitionmcmc} which sample DAGs according to their posterior distribution. Hybrid approaches combine constraint-based and Bayesian methods to improve the speed and accuracy of the sampling procedure \citep{Tsamardinos2006, efficientsampling}. 

One of the first approaches attempting to incorporate uncertainty about the graphical structure for causal inference was the Intervention-calculus when the DAG is Absent (IDA) algorithm of \cite{ida}. This approach, restricted to jointly Gaussian distributions, estimates an equivalence class via the PC algorithm and enumerates the possible causal effects that are entailed by the estimated equivalence class. More sophisticated approaches for estimating intervention distributions without prior causal graph knowledge rely on discrete or Gaussian-linear models combined with Bayesian structure learning algorithms \citep{moffa2017, viinikka2020, Castelletti2021}. Such models fully take the causal graph uncertainty into account but in the continuous case are limited to linear models, which can be unsuitable for many realistic applications \citep{campos2006, Zhang2014, kontio2020}.

Non-linear additive models provide a way of representing causal relationships with more general joint distributions beyond the Gaussian-linear case. Identifying the DAG for non-linear, additive models has been studied extensively \citep{Hoyer2008, Peters2014} and related causal inference approaches have been developed \citep{CAM_2014, marg_int}. Such methods yield a single causal estimate and therefore do not provide the means to quantify the uncertainty of the estimate due to lack of prior information on the structure or the parameters. Recently, an interesting approach by \cite{geffner2022} uses advances in differentiable causal discovery by \cite{NoTears} to estimate non-linear intervention effects from observational data. The approach is however not Bayesian since the posterior over DAGs is approximated by a variational distribution which models the probability of each edge as an independent Bernoulli random variable. 

In this work, we focus on the problem of causal inference in the Gaussian Process Network (GPN) model of \cite{GPnetworks}, which uses Gaussian processes to allow flexible, non-linear relationships between variables. We develop methods to estimate intervention distributions adopting the Bayesian approach of treating all unknown parameters as random. The result is a streamlined procedure that takes into full account the uncertainty of the graph and parameters. We consider two separate approaches to deriving causal quantities: a Monte Carlo (MC) and a ``local'' approximation, based respectively on truncated factorization and adjusting for variables satisfying the backdoor criterion \citep{Pearlmut}. Although these two approaches are equally well-grounded in causal inference theory, in practice they require different modelling assumptions due to the non-linear nature of GPNs. After a brief review of relevant background material in section \ref{sec:bac}, we treat the two aforementioned causal inference approaches respectively in sections \ref{sec:glob} and \ref{sec:loc}. Section \ref{sec:exps} is instead devoted to implementation details and practical examples.

\section{Background}
\label{sec:bac}
\subsection{Causal Bayesian Networks}
Directed Acyclic Graphs (DAGs) are a convenient tool for representing the causal relationships among a set of random variables $\textbf{X} = \{X_1,...\,,X_n\}$. A variable $X_i$ has an edge directed towards $X_j$ if intervening on $X_i$ affects the distribution of $X_j$ when all the other variables $X_{\setminus ij}$ are fixed, i.e.~there exists $x \neq x'$ such that
\begin{equation}
    p(X_j\given\textrm{do}(X_i=x,X_{\setminus ij}=\textbf{k}) \,\neq\, 
    p(X_j\given\textrm{do}(X_i=x',X_{\setminus ij}=\textbf{k})
\end{equation}
for every $\textbf{k}$ \citep{mooij2016}. The DAG representation provides a compact, graphical abstraction of the causal relations underlying the data-generating process \citep{bnop}. Causal Bayesian network models are defined as a pair $\langle p(\mathbf{X}),\mathcal{G} \rangle$ in which the joint distribution is Markov with respect to the causal DAG $\mathcal{G}$, i.e.
\begin{equation}
\label{Markov}
    p(\mathbf{X}) \,=\, \prod_{i=1}^n \, p(X_i\given\textrm{Pa}^{\mathcal{G}}_{X_i})
\end{equation}
where the parents $\textrm{Pa}^{\mathcal{G}}_{X_i}$ are the set of variables that have an outgoing edge directed towards $X_i$ in the causal graph $\mathcal{G}$. Causal Bayesian networks require the assumption that the model is causally sufficient, meaning that all common causes of any two variables are themselves accounted for in the graph \citep{bewaredag}.

Equation (\ref{Markov}) equivalently states that every variable is independent of its non-effects given its parents \citep{lauritzen1996}. This property, known as the causal Markov condition, provides the link between the causal statements contained in the DAG and the factorization of the variables' joint distribution. The causal Markov condition follows the intuition that the dependence between each cause and effect is mediated by the direct causes (the parents) of the effect. Conditioning on a variable's parents therefore renders it independent from all variables other than its effects.

\subsection{Gaussian Processes}
\label{sec:GP_recap}
Gaussian Processes (GPs) are popular tools for regression and classification that have been used extensively in the statistics and machine learning literature \citep{MacKay2003, GPOP}. A GP is defined as a (potentially infinite) collection of random variables $\{f(x_i)\}_{x_i \in \mathcal{X}}$ such that for every finite subset $\{x_1,...\,,x_k\}$ of the index set $\mathcal{X}$, the marginal distribution over that subset has a multivariate Gaussian distribution:
\begin{equation}
\label{GPnorm}
    \{f(x_1),...\,,f(x_k)\} \,\sim\, \mathcal{N}(\bm{\mu}, \Sigma).
\end{equation}
GPs are used to describe distributions over functions, where the random variables represent the values of the function $f(x)$ at the location $x$ and the index set $\mathcal{X}$ represents the set of possible inputs. A GP is said to be centered when its expectation is zero, i.e.~$\mathbb{E}[f(x_i)] = 0 \,~\forall\, x_i \in \mathcal{X}$. In such a case the mean $\bm{\mu}$ in (\ref{GPnorm}) is zero and the process is fully described by the matrix $\Sigma$, whose entries are determined by the covariance function $k(x_i,x_j) = \textrm{Cov}[f(x_i), f(x_j)] \,~\forall\, x_i, x_j \in \mathcal{X}$. 

Centered GPs are commonly used as priors for relating inputs $x$ to realizations of an outcome variable $Y$ in combination with a Gaussian likelihood:
\begin{equation}
\begin{aligned}
    f(x) &\,\sim\, \mathcal{GP}\left(0, k(.\,,.)\right) \\
    Y\given f(x) &\,\sim\, \mathcal{N}\!\left(f(x),\sigma^2\right).
\end{aligned}
\end{equation}
The marginal likelihood of the observed realizations $y$ of the outcome variable is given by the Gaussian likelihood
\begin{equation}
\label{GPlik}
    y \,\sim\, \mathcal{N}(0, K+\sigma^2I)
\end{equation}
where $K$ represents the $N \times N$ Gram matrix
\begin{equation*}
    K_{ij} = k(x_i,x_j)\,.
\end{equation*}
Posterior inference on $f$ is performed by exploiting the fact that the realizations $y$ of the outcome variable and the GP $f^*$ evaluated at the test points $x^*$ both have a Gaussian distribution:
\begin{equation}
    \begin{pmatrix} y \\ f^* \end{pmatrix} \sim\,
    \mathcal{N} \left(0 \,,\,
    \begin{bmatrix} 
        K+\sigma^2I & K^* \\
        K^{*\top} & K^{**} 
    \end{bmatrix}\right)
\end{equation}
where the Gram matrices $K^*$ and $K^{**}$ are given by 
\begin{equation*}
    K^*_{ij} = k(x_i,x^*_j)\,, \quad~ K^{**}_{ij} = k(x^*_i,x^*_j)\,.
\end{equation*}
The posterior GP of $f^*$ is then derived as the conditional of a Gaussian distribution:
\begin{equation}
\label{GPpost}
    f^*|\, y \,\sim\, \mathcal{N} \left(K^{*\top} (K+\sigma^2I)^{-1} y,\,
    K^{**} - K^{*\top} (K+\sigma^2I)^{-1} K^*\right).
\end{equation}

\subsection{Gaussian Process Networks}
Bayesian networks whose conditional distributions are modeled via GP priors were first introduced by \cite{GPnetworks} as GPNs. The following structural equation model for a generic variable $X_i$ for $i\in\{1,...\,,n\}$ describes a GPN:
\begin{equation}
\begin{aligned}
\label{strucmodel}
    X_i &\,=\, f_i(\textrm{Pa}_{X_i}) + \varepsilon_i \\
    f_i &\,\sim\, \,\mathcal{GP}(0, k_i(.\,,.)) \\
    \varepsilon_i &\,\sim\, \mathcal{N}(0, \sigma_i^2)
\end{aligned}
\end{equation}
where the Gaussian noise variables $\varepsilon_i$ are independent of the data. Thanks to the nonparametric nature of GPs, the above model leads to a highly flexible class of conditional distributions $p(X_i\given \textrm{Pa}_{X_i})$ for continuous data. GPNs therefore serve as suitable models when the relationships among the variables in the network are unknown and the user seeks to avoid making strict a priori assumptions on the conditional distributions.

Each kernel function $k_i(.\,,.)$ is parameterized by a set of unknown parameters $\bm{\theta}_i$ which together with the noise variance $\sigma_i^2$ form the hyperparameter set $\Theta_i$ of the conditional distribution $p(X_i\given \textrm{Pa}_{X_i})$. 
When performing inference on the GPN, these hyperparameters must be determined from the available data. The approach of \cite{GPnetworks} is to select the maximum a posteriori (MAP) hyperparameters $\hat{\Theta}_i = \{\hat{\bm{\theta}}_i, \hat{\sigma}_i^2\}$ that maximize the product between the likelihood (\ref{GPlik}) of the observations $x_i$ and the prior $p(\Theta_i\given \textrm{Pa}_{X_i})$:
\begin{equation}
    \hat{\Theta}_i \,=\, \argmax_{\Theta_i}{\,p(x_i\given \textrm{Pa}_{X_i}, \Theta_i) \, p(\Theta_i\given \textrm{Pa}_{X_i})}\,.
\end{equation}
The prior depends on the parent set due to the dimension of the kernel function's parameters $\bm{\theta}_i$ typically increasing with the size of the parent set.
Inference on other features of the network is then performed by plugging the obtained hyperparameter values $\hat{\Theta}_i$ into the conditional distributions. For example, inference on the function $f_i^*$ evaluated at some generic test points of the parents is performed by plugging $\hat{\Theta}_i$ into equation (\ref{GPpost}) obtaining the posterior $p(f_i^* \given x_i, \hat{\Theta}_i)$.

A fully Bayesian approach, on the other hand, would require integrating the posterior of interest (\ref{GPpost}) over the whole posterior distribution of the hyperparameters:
\begin{equation}
    p(f_i^* \given x_i) \,=\, \int p(f_i^* \given x_i, \Theta_i)\,p(\Theta_i \given \textrm{Pa}_{X_i}, x_i)\,d\Theta.
\end{equation}
Although computationally more expensive, the Bayesian approach of integrating over the posterior allows us to fully take into account the uncertainty regarding the values of the hyperparameters. In practice, the posterior $p(\Theta_i \given x_i)$ is usually estimated via MCMC methods \citep{GPmcmc}.

Due to the difficulties in estimating marginal likelihoods, Bayesian structure inference for GPNs is particularly complex. Recently, \cite{giudice2023bayesian} introduced an importance sampling-based approach to sample from a GPN's posterior over DAGs, which allows a fully Bayesian treatment of the hyperparameters.

\section{Causal Inference with Gaussian Process Networks}
\label{sec:glob}
In this section, we describe a procedure to perform posterior inference on a generic intervention distribution for a given GPN $\langle p_{\mathcal{G}}(\mathbf{X}),\mathcal{G} \rangle$ starting from the truncated Markov factorization. We will treat two cases separately according to whether the DAG $\mathcal{G}$ is known or not. In the latter case, the Bayesian approach involves integrating the intervention effect of interest over the posterior distribution of $\mathcal{G}$. In section \ref{sec:nodag_glob} we will therefore combine our GPN causal inference procedure with an MCMC scheme to sample an ensemble of DAGs. The procedure results in a Monte Carlo estimate of the posterior intervention distribution which fully takes into account the uncertainty regarding both the structure and parameters.

When tackling causal inference tasks, a general quantity of interest is $p(Y\given\textrm{do}(X=x))$, i.e.~the distribution of an outcome variable of interest $Y \in \textbf{X}$ for a given intervention on another variable $X \in \textbf{X} \setminus Y$. One is typically interested in evaluating this function at a range of values $x$, which correspond to different interventions on the variable $X$. Interventions can also be considered on larger sets of variables. The intervention distribution can be derived using the truncated Markov factorization and the related concept of an \textit{interventional}, or \textit{manipulated} network \citep{Pearlmut}. In such a network all edges incoming to the intervention variable $X$ are deleted and the joint distribution is ``truncated'' by constraining $X$ to the intervention value. All the conditional distributions therefore remain identical to the original BN except for the distribution of $X$ which has all of its mass concentrated on the intervention value $x$. Formally, for a given causal BN $\langle p_{\mathcal{G}}(\mathbf{X}),\mathcal{G} \rangle$, the interventional network $\langle p_{\mathcal{H}}(\mathbf{X}),\mathcal{H} \rangle$ is such that
\begin{itemize}
    \item $\textrm{Pa}^{\mathcal{H}}_{X} = \emptyset$ and $p_{\mathcal{H}}(X=x)=1$.
    \item For all other $X_i \in \textbf{X}\setminus X$,\, $p_\mathcal{H}(X_i\given\textrm{Pa}^{\mathcal{H}}_{X_i}) =
    p_\mathcal{G}(X_i\given\textrm{Pa}^{\mathcal{G}}_{X_i})$.
\end{itemize}
The target quantity of interest can then be simply written as a marginal distribution in the interventional network:
\begin{equation}
    p(Y\given\textrm{do}(X=x)) \,=\, p_{\mathcal{H}}(Y)\,.  
\end{equation}

An equivalent derivation of the above target is given by the backdoor adjustment formula. For any set $Z$ that satisfies the backdoor criterion \citep{Pearl1993}, the intervention distribution can be written as
\begin{equation}
    p(Y\given\textrm{do}(X=x)) \,= \int p_{\mathcal{G}}(Y\given X=x, Z)\,dP(Z)\,.
\end{equation}
Importantly, the set $\textrm{Pa}_X$ of parents of $X$ is guaranteed to satisfy the backdoor criterion for the effect of $X$ on $Y$, although it may not be the most efficient \citep{Brookhart06,effadj}. Indeed, different adjustment sets that satisfy the backdoor criterion yield equally consistent estimates but the resulting variance can differ in finite samples \citep{perkovic2018, KuipersMoffav}. In the sections that follow we adjust according to the parent set, but the methods illustrated in this work are easily extendable to more efficient sets \citep{betterset1,betterset2}.
If $Y \notin \textrm{Pa}_X$ then
\begin{equation}
\label{target}
    p(Y\given\textrm{do}(X=x)) \,= \int p_{\mathcal{G}}(Y\given X=x, \textrm{Pa}_X)\,dP(\textrm{Pa}_X).
\end{equation}
If $Y$ instead belongs to the parent set of $X$ then $p(Y\given\textrm{do}(X=x)) = p_{\mathcal{G}}(Y)$.

The above equation formulates the target as a function of the intervention, outcome and adjustment variables only, which can be modeled directly to obtain an approximation of the intervention distribution. In the rest of this section we focus on sampling from the posterior intervention distribution of the intervention distributions, while approximations based on the adjustment formula are deferred to section \ref{sec:loc}.

\subsection{Causal Inference for a Given DAG}
\label{sec:dag_glob}
Let all hyperparameter sets $\Theta_i ~\forall \,i\in\{1,...\,,n\}$ of the model (\ref{strucmodel}) be known for the time being. The marginal distribution $p_{\mathcal{H}}(Y)$ of the interventional network is however not available in closed form for the GPN model; we resort therefore to a Monte Carlo (MC) approximation. The MC approximation is obtained by propagating samples following the truncated Markov factorization of the BN model. Without loss of generality, assume that the topological ordering of the nodes follows the indices $1,...\,,n$; we can then sample from the intervention distribution by propagating $M$ values of $x$ down the network according to the topological ordering of the nodes in the interventional graph $\mathcal{H}$. The values $x_i$ of a generic node are sampled from the conditional distribution of $X_i$ given the previously sampled values of its parents:
\begin{equation}
\begin{aligned}
\label{propagate}
    x_{1,m} &\sim p_{\mathcal{H}}(X_1) \\
    x_{2,m} &\sim p_{\mathcal{H}}(X_2\given\textrm{Pa}^{\mathcal{H}}_{X_2},x_{1,m})  \\
    &~\,\vdots  \\
    x_{n,m} &\sim p_{\mathcal{H}}(X_n\given\textrm{Pa}^{\mathcal{H}}_{X_n}, x_{n-1,m},...\,,x_{1,m})\,. 
\end{aligned}
\end{equation}
The conditional distributions $p_{\mathcal{H}}(X_i\given \textrm{Pa}_{X_i})$ are given by the structural equation model (\ref{strucmodel}), where the functions $f_i$ are sampled from the posterior (\ref{GPpost}). The multivariate Gaussian posterior (\ref{GPpost}) lends itself to sampling a range of values from the conditional distributions at once, leading to an efficient, vectorizable procedure to obtain the different samples corresponding to different levels of the intervention variable. Furthermore, for each intervention $\textrm{do}(X=x)$, the procedure allows one to obtain samples from the intervention distribution $p(Y\given\textrm{do}(X=x))$ for all variables $Y$ downstream from $X$ in the topological order. Algorithm 1 shows pseudo-code for sampling from the intervention distributions $p(Y\given\textrm{do}(X=x))$ for all ordered pairs of variables $X,Y.$

\begin{algorithm}
\caption{Causal Inference for a Known GPN}
\hspace*{\algorithmicindent} \textbf{Input} Graph $\mathcal{G}$, conditional distributions $p_\mathcal{G}(X_i\given \textrm{Pa}^{\mathcal{G}}_{X_i})~\forall\, i = 1,...\,,n$\\
\hspace*{\algorithmicindent} \textbf{Output} Samples from $p(X_\ell\given\textrm{do}(X_k=x_k))~\forall\, k,\ell = 1,...\,,n,\, k \neq \ell$
\begin{algorithmic}[1]
\State Obtain topological order $\mathcal{T}$ of the variables $\textbf{X}$ in $\mathcal{G}$.
\For{$k \in \{1,...\,,n\}$}
    \State Build interventional network $\mathcal{H}_k$, with $X_k = x_k$.
    \For{$\ell \in \mathcal{T}$}
        \For{$m \in \{1,...\,,M\}$}
            \State Sample $x_{\ell,m} \sim p_{\mathcal{H}_k}(X_\ell\given\textrm{Pa}^{\mathcal{H}_k}_{X_\ell},\{x_j\}_{j<k})$. 
        \EndFor
        \State Save $x_{\ell,m}$ as samples from $p(X_\ell\given\textrm{do}(X_k=x_k))$.
    \EndFor
\EndFor
\end{algorithmic}
\end{algorithm}
 
When the hyperparameters of the BN are unknown they must be estimated from the data. In the GPN case the hyperparameter set $\Theta_i$ for each conditional distribution $p_{\mathcal{H}}(X_i\given \textrm{Pa}_{X_i})$ includes the hyperparameters $\theta_i$ of the kernel function of the GP prior on $f_i$, as well as the variance $\sigma^2_i$ of the Gaussian noise $\varepsilon_i$. Sampling from the conditional distributions of equation (\ref{propagate}) then requires the additional step of sampling from the hyperparameters' posterior distribution. Although this distribution is not available in closed form, MCMC methods can efficiently provide samples from the posterior \citep{GPmcmc}. Conditionally on these values, we sample a realisation from the GP posterior distribution at the location of its parents' sampled values according to equation (\ref{GPpost}). Finally, we add a sample from the independent Gaussian noise $\varepsilon_i$:
\begin{equation}
\begin{aligned}
\label{samplin}
    \theta_{i,m}, \sigma^2_{i,m} &\,\sim\, p_{\mathcal{H}}(\Theta_i\given D)  \\
    f_{i,m} &\,\sim\, p(f_i\given D, \theta_{i,m})  \\
    \varepsilon_{i,m} &\,\sim\, \mathcal{N}(0, \sigma^2_{i,m})  \\
    x_{i,m} &\,=\, f_{i,m} + \varepsilon_{i,m}\,. 
\end{aligned}
\end{equation}
The resulting samples $x_{i,m}$ are then samples from the posterior predictive distribution $p(X_i\given \textrm{Pa}_{X_i}, D)$. These can be integrated into line $6$ of algorithm 1 to obtain samples from the (posterior predictive) intervention distribution $p(X_\ell\given\textrm{do}(X_k=x_k), D)$.

If the target quantity of interest is the expectation $\mathbb{E}(X_\ell\given\textrm{do}(X_k=x_k))$, this is estimated directly with the samples $f_{\ell,m}$ from the posterior $p_{\mathcal{H}_k}(f_\ell\given\textrm{Pa}^{\mathcal{H}_k}_{X_\ell},\{x_j\}_{j<\ell})$, omitting the additive noise when sampling from the conditional distribution of $Y$ given its parents.

\subsection{Without a Known DAG}
\label{sec:nodag_glob}
In the absence of a known graphical structure $\mathcal{G}$, we must resort to a structure learning algorithm to estimate the underlying DAG. Bayesian methods consist of an MCMC scheme in the DAG space to obtain samples from the posterior distribution of DAGs. Such samples from the posterior can then be used to estimate the true posterior distribution of any feature of interest $\Psi$ in the BN model via MC integration:
\begin{equation}
\label{MCintegration}
    p(\Psi\,\vert\, D) \,=\, \frac{1}{M} \sum_{m=1}^M \,p(\Psi\,\vert\,\mathcal{G}_m)\,,
    \,\quad\,\mathcal{G}_m \,\sim\, p(\mathcal{G}\given D)
\end{equation}
where $M$ denotes the number of samples obtained via the MCMC algorithm. Sampling graphs from GPNs is however a computationally expensive procedure due to the requirement of integrating the marginal likelihood with respect to the prior distribution over the hyperparameters \citep{BeingBayesian}. The approach of \cite{giudice2023bayesian} uses an approximation of the posterior $q(\mathcal{G}\given D)$ to obtain samples $\mathcal{G}_1,...\,,\mathcal{G}_M$ of DAGs together with a set of weights $w_1,...\,,w_M$ which can be used to make inference on the true posterior via importance sampling. 

By applying algorithm 1 to each sampled DAG $\mathcal{G}_m$, we can compute and save a sample $\mathcal{V}_m$ from the intervention distribution of interest conditionally on the sampled DAG. The collected samples $\mathcal{V}_1,...\,,\mathcal{V}_M$ together with the importance weights $w_1,...\,,w_M$ can then be used to make full posterior inference, for example on the expectation $\mathbb{E}(X_\ell\given\textrm{do}(X_k=x_k))$:
\begin{equation}
\label{totalexpectation}
    \mathbb{E}(X_\ell\given\textrm{do}(X_k=x_k)) \,\approx\, \frac{\sum_{m=1}^M \,\mathcal{V}_m w_m}{\sum_{m=1}^M w_m}\,,~~~
     \mathcal{V}_m \,\sim\, p(X_\ell \given \textrm{do}(X_k=x_k), \mathcal{G}_m)\,,~~~ \mathcal{G}_m \,\sim\, q(\mathcal{G}\given D)\,.
\end{equation}
Besides inference on the posterior mean, the approach also allows computing measures of uncertainty such as standard deviations or quantiles.

Computing the importance weights $w_1,...\,,w_M$ requires learning all the conditional distributions of each variable given their parents in the sampled DAGs. These can be saved during the graph sampling process and then directly passed on to algorithm 1 to avoid being re-computed during the MC inference step. The procedure is described as pseudocode in algorithm $2$.

\begin{algorithm}
\caption{Causal Inference in GPNs Without a Known DAG}
\hspace*{\algorithmicindent} \textbf{Input} Data $D$ of $n$ variables \\
\hspace*{\algorithmicindent} \textbf{Output} Estimates of $\mathbb{E}(X_\ell\given\textrm{do}(X_k=x_k))~\forall\, k,\ell = 1,...\,,n,\, k \neq \ell$
\begin{algorithmic}[1]
\For{$m \in \{1,...\,,M\}$}
    \State Sample DAG $\mathcal{G}_m$ and its weight $w_m$ according to the posterior $p(\mathcal{G}_m\given D)$.
    \State Save all conditional distributions $p(X_i\given \textrm{Pa}^{\mathcal{G}_m}_{X_i}),~\forall\, i = 1,...\,,n$.
    \State Sample $x_{\ell,m}$ with algorithm~1 for given $\mathcal{G}_m$ and $p(X_i\given \textrm{Pa}^{\mathcal{G}_m}_{X_i})~
    \forall\, i,\ell = 1,...\,,n$. \Comment{Alg.~(1)}
\EndFor
\State Compute $\mathbb{E}(X_\ell\given\textrm{do}(X_k=x_k))$ via importance sampling $\forall\, k,\ell = 1,...\,,n,\, k \neq \ell$. \Comment{Eq.~(\ref{totalexpectation})}
\end{algorithmic}
\end{algorithm}

\section{Local approximation}
\label{sec:loc}
As the dimension of the network increases, the procedure of sampling from the truncated factorization outlined in algorithm 1 becomes increasingly expensive. Every sampling step requires learning the hyperparameters of the conditional distributions, which becomes costlier as the paths between the intervention variable $X$ and the outcome variable $Y$ increase in number and length.
To address this issue, in this section, we follow a different approach which estimates conditional probabilities based on \textit{local} variables only, e.g.~we only use partial information of the graph \citep{marg_int}. This is opposed to the approach described in section \ref{sec:glob}, which relies \textit{globally} on all variables in the graph to estimate an intervention distribution. 

The adjustment formula (\ref{target}) is a useful tool for our purposes since it allows us to express the intervention distribution as a function of $X$, $Y$ and $\textrm{Pa}_X$ only. This avoids the potentially long chains of sampling statements in equation (\ref{propagate}) and allows computing at once intervention distributions for a large range of interventions $\textrm{do}(X=x)$.

The approach requires modeling the joint relationship of $Y$ given $X$ and its parent set. The reduced computational effort therefore comes at the cost of estimating the conditional distribution $ p(Y\given X=x, \textrm{Pa}_X)$ with a single GP regression model. The quality of the resulting estimate will therefore depend on the ability of the GP to capture the relationships between $X$ and $Y$ as well as between $Y$ and the parent set. Different modeling choices for the conditional distribution can therefore lead to different estimates.

An additive GP model provides a simple, natural starting point since it allows us to isolate the different contributions of each variable and to make inference separately on the components of the mean:
\begin{equation}
\begin{aligned}
\label{full_addmodel}
    Y &\,=\, f(X) + \!\sum_{Z \in \textrm{Pa}_X} \!g_Z(Z) + \varepsilon \\
    f &\,\sim\, \,\mathcal{GP}\left(0,\, k_X(x,x')\right)  \\
    g_Z &\,\sim\, \,\mathcal{GP}\left(0,\, k_Z(z,z')\right)  \\
    \varepsilon &\,\sim\, \mathcal{N}(0, \sigma^2). 
\end{aligned}
\end{equation}
For identifiability, the means $\mathbb{E} g_Z(Z)$ are assumed to be zero \citep{additive_marg, CAM_2014}. The intervention distribution of interest (\ref{target}) then corresponds to
\begin{equation}
    Y\given\textrm{do}(X=x) \,\sim\, \mathcal{N}\left( f(x) + \!\sum_{Z \in \textrm{Pa}_X}\! \mathbb{E} \,g_Z(Z)\,,\,\sigma^2\right). 
\end{equation}
Since in practice the constant term $\sum \mathbb{E} g_Z(Z)$ can be discarded due to the zero mean requirement, causal inference in such a model boils down to marginal inference on the quantity $f(x)$ for the mean and $\sigma^2$ for the variance. The hyperparameters $\sigma^2$ and those of the kernel functions of $f$ and $g_Z$ can be learned by either maximizing the marginal likelihood or sampling from the hyperparameters' posterior distribution. Conditionally on the hyperparameter values, posterior inference on the component $f$ is available in closed form. To see this, we first note that an additive model in separate components gives rise to an additive kernel structure that follows the same decomposition \citep{additiveGPs}. We can then write our additive model as
\begin{equation}
\label{part_addmodel}
    Y \,=\, f(X) + g(\textrm{Pa}_X) + \varepsilon 
\end{equation} 
with
\begin{equation}
    g \,\sim\, \,\mathcal{GP}\left(0,\, \sum_{Z \in \textrm{Pa}_X}\!k_Z(z,z')\right).
\end{equation}
If the GPs in (\ref{full_addmodel}) are a priori independent, then we can apply the same procedure in section \ref{sec:GP_recap} to derive the marginal posterior distribution of $f$. The full joint distribution of the realizations $y$ of the outcome variable $Y$ and the function values $f^*$ and  $g^*$ at the test locations is \citep{duvenaud_thesis}
\begin{equation}
    \begin{bmatrix} y \\ f^* \\ g^* \end{bmatrix} \,\sim\,
    \mathcal{N} \left(0 \,,\,
    \begin{bmatrix} K_X + \sum K_Z + \sigma^2I & K_X^* & K_Z^* \\ 
                    K_X^{*\top} & K_X^{**} & 0 \\
                    K_Z^{*\top} & 0 & K_Z^{**} \end{bmatrix}\right).
\end{equation}
The gram matrix notation follows that of section \ref{sec:GP_recap}. The posterior of interest is then
\begin{equation}
\label{marg_post}
    f^*|\,y \,\sim\, \mathcal{N} \left(K_X^{*\top}\!\left(K_X + \sum K_Z + \sigma^2I\right)^{-1}\!y \,,\,\, K_X^{**} - K_X^{*\top}\!\left(K_X + \sum K_Z + \sigma^2I\right)^{-1}\!K_X^*\right).
\end{equation}
The above equation allows us to perform efficient posterior inference on $\mathbb{E}(Y|\textrm{do}(X=x))$ at a set of test locations $x$ without the need for learning additional hyperparameters or sampling steps. 

The downside of modeling conditional distributions such as $p(Y\given X=x, \textrm{Pa}_X)$ locally is that it paves the way for potential misspecification in the original model (\ref{strucmodel}). In a GPN model, conditional distributions are generally complex due to the compounding of non-linear parent-child relations; the MC approach outlined in section \ref{sec:dag_glob} accounts for this by learning all necessary parent-child relations. A local approximation on the other hand relies on a single model to jointly learn the relationship between $Y$ and $X$ as well as the dependencies between $X$ and its parent set. Simulation studies in section \ref{sec:exps} show that the flexible nature of GPs is generally able to provide reasonable approximations for the conditional distributions, and although the local approximation does not match the exact posterior, its use can still lead to adequate estimates for causal quantities of interest.

The hyperparameter set $\Theta$ of equation (\ref{marg_post}) contains the noise variance $\sigma^2$, as well as all the parameters of the covariance functions  $k_X$ and $k_Z$, for all $Z \in \textrm{Pa}_X$. When $\Theta$ is unknown, the Bayesian approach involves sampling from its posterior according to model (\ref{full_addmodel}). Samples from $f^*|\,y$ can then be taken conditionally on the sampled hyperparameters:
\begin{equation}
\begin{aligned}
    \Theta_m &\,\sim\, p(\Theta\given D)  \\
    f^*_{m}|\,y &\,\sim\, p(f^*\given y, \Theta_{m})\,.
\end{aligned}
\end{equation}

\subsection{Local Approximation Without a Known DAG}
In the absence of a known DAG, we take structure uncertainty into account by integrating the quantity of interest over the posterior distribution of DAGs $p(\mathcal{G}\given D)$. As discussed in section \ref{sec:nodag_glob}, MCMC schemes can provide samples from such a posterior. Once a sufficiently large number $M$ of samples have been obtained, we need to compute for each sampled DAG $\mathcal{G}_i$ the quantity of interest, for example, the posterior $p(f^*|\,y, \mathcal{G}_i)$ for inference on the expectation $\mathbb{E}(Y\given\textrm{do}(X=x))$. 

Computing $M$ times the posterior $f^*|\,y$ in equation (\ref{marg_post}) is however computationally expensive, since it requires learning again the hyperparameters of the GP regression at every iteration. To avoid redundant optimizations and matrix inversions, we note that $f^*|\,y$ depends on the graph $\mathcal{G}_i$ only through the parent set $\textrm{Pa}_X$. Let $\Omega_X$ be a random variable denoting the nodes of the parent set of a variable $X$ taking values in the power set of the nodes of $\textbf{X} \setminus \{X,Y\}$ and let $S_j$ be a generic element of such a power set. We can then perform the MC integration over the different parent sets instead of over all sampled DAGs: 
\begin{equation}
\label{fx_mixture}
    p(f^*|\, y) \,\approx\, \frac{1}{M}
    \sum_{j=1}^M p(f^*|\, y, \Omega_X= S_j) \,,\,\quad
    S_j \,\sim\, p(\Omega_X \given D)\,.
\end{equation}
In theory, the number of all possible parent sets grows exponentially with the number of variables; in practice however, we expect the number of sampled parent sets to be much smaller. This is because in non-pathological cases the posterior will be peaked around the true parent set, giving nearly zero probability mass to most parent configurations. 

We can estimate the posterior distribution over parent sets $p(\Omega_X\given D)$ following the MC integration method (\ref{MCintegration}). As mentioned in section \ref{sec:nodag_glob}, it is more efficient to sample GPNs from an approximate posterior $q(\mathcal{G} \given D)$ together with a set of importance weights \citep{giudice2023bayesian}. The estimated distribution over parent sets must then be weighted in the same way as in equation (\ref{totalexpectation}):
\begin{equation}
\label{Pa_dis_imp}
    p(\Omega_X= S_j\given D) \,\approx\, \frac{
    \sum_{i=1}^M w_i\, \scalebox{1.1}{$\boldone$}\scalebox{0.8}{$(\mathcal{P}a^{\mathcal{G}_i}_X = S_j)$}}{\sum_{i=1}^M w_i} \,,
    \,\quad\, \mathcal{G}_i \,\sim\, q(\mathcal{G}\given D)\,.
\end{equation}
The resulting estimate (\ref{fx_mixture}) is therefore a mixture of GPs, weighted by the relative frequency of each parent set in the sampled DAGs. In practice, we can simply sample for each parent set $S_j$ a number of posterior samples of $f^*$ proportional to its posterior probability $p(\Omega_X=S_j \given D)$. The whole procedure is summarized in algorithm $3$. 

\begin{algorithm}
\begin{algorithmic}[1]
\caption{Local Approximation Without a Known DAG}
\For{$i \in \{1,...\,,M\}$}
    \State Sample DAG $\mathcal{G}_i$ and its weight $w_i$ according to the posterior $p(\mathcal{G}_i\given D)$.
\EndFor
\For{every ordered pair of variables $X,Y$ we wish to compute $\mathbb{E}(Y\given \textrm{do}(X=x))$}
    \State Estimate the posterior distribution of the parent sets $\Omega_X$ of $X$. \Comment{Eq.~(\ref{Pa_dis_imp})}
    \For{every sampled parent set $\Omega_X= S_j$}
        \State Fit additive GP regression of $Y$ on $X$ and $S_j$:
        $$Y = f_j(X) + \sum_{k=1}^{|S_j|}g_{j,k}(S_{j,k}) + \varepsilon_j.$$
        \State Save samples $\bm{f}_j^*$ from $p(f^*|\, y, \Omega_X= S_j)$ in number proportional to
        
        \quad\quad $p(\Omega_X= S_j\given D)$. \Comment{Eq.~(\ref{marg_post})}
    \EndFor
    \State Estimate $\mathbb{E}(Y\given\textrm{do}(X=x))$ as a mixture $\{\bm{f}_j^*\}_j$. \Comment{Eq.~(\ref{fx_mixture})}
\EndFor
\end{algorithmic}
\end{algorithm}


\section{Experimental Results}
\label{sec:exps}
In this section we provide results and discuss implementation details concerning the previously described algorithms to derive intervention distributions in GPNs. For all of our experiments we model each GP prior in the structural equation model (\ref{strucmodel}) with an additive squared exponential kernel:
\begin{equation}
\label{addkernel}
    k(.\,,.) \,= \sum_{j=1}^{|\textrm{Pa}_{X}|} \textrm{exp}\left(-\frac{||x_j-x_j'||^2}{2\theta_j^2}\right)
\end{equation}
where each lengthscale $\theta_j$ controls the rate of decay of the correlation in terms of distance between two samples of the $j$th parents. Although GPNs can model a wider range of relationships between multiple inputs and an output, we choose the above additive model as it has an established history in the machine learning literature \citep{elements} and has proven able to satisfactorily approximate more complex functional relationships \citep{marg_int}. Following the original definition of GPNs \citep{GPnetworks}, we assign marginal Gaussian distributions to nodes without parents. 

\begin{figure}[h]
\centering
\includegraphics[width = 3.5cm]{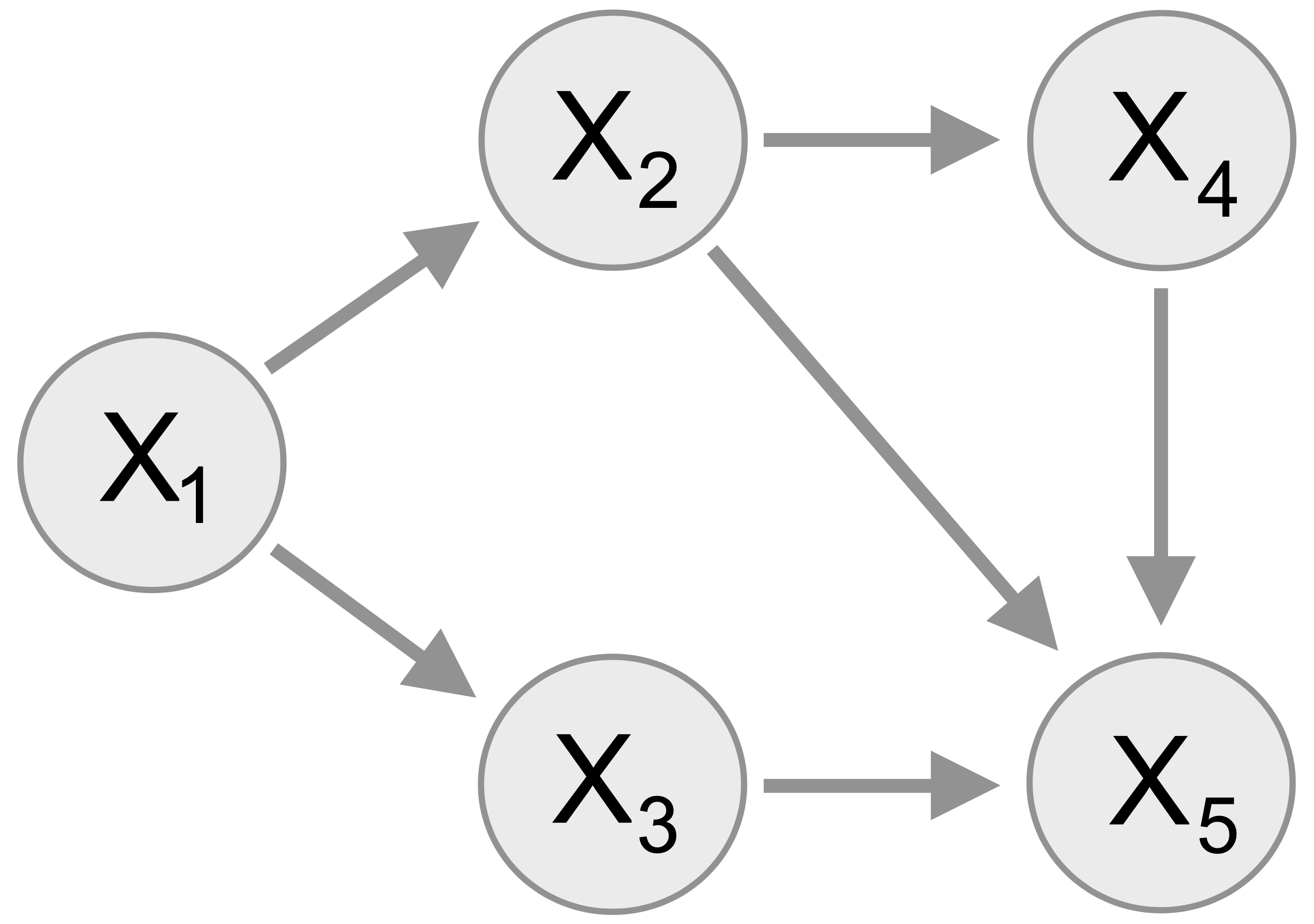} 
\caption{The DAG used to generate the data for figures \ref{fig:glob_dag}-\ref{fig:loc_nodag}.}
\label{fig:dag}
\end{figure}

To provide an example of the implementation of the algorithm based on the truncated Markov factorization and its ``local'' approximation proposed respectively in sections \ref{sec:glob} and \ref{sec:loc}, we generate synthetic data from a known GPN with $n=5$ nodes. For the graphical structure we use the DAG in figure \ref{fig:dag}. The functional relationships between each variable and its parents are generated as a weighted combination of different Fourier components:
\begin{equation} 
\label{Fourier}
    X_i \,= \sum_{j \mid X_{j}\, \in \,\textrm{Pa}_{X_i}}\!\beta_{i,j}\Bigg\{\,u_{i,j,0}\,X_j\,+\,\sum_{k=1}^6 \Big[ \, v_{i,j,k} \sin{(k X_j)} \,+\, u_{i,j,k} \cos{(k X_j)}\Big]\Bigg\} \,+\, \epsilon_i.
\end{equation}
The noise term $\varepsilon_i$ is sampled from a $\mathcal{N}(0,\frac{1}{2})$ distribution, while the weights $v_i$ and $u_i$ are sampled from a $\textrm{Dir}(\alpha_0,...\,,\alpha_6)$ distribution with $\alpha_k = e^{-k}$. The edge weights $\beta_i$ are instead sampled from a uniform distribution on $(-2, -\frac{1}{2}) \cup (\frac{1}{2},2)$. Equation (\ref{Fourier}) provides data from a given DAG with random, non-linear functional relationships for every edge in the network.

For each variable, $50$ samples are then generated according to equation (\ref{Fourier}). We employ the hybrid version of the partition MCMC algorithm \citep{partitionmcmc, efficientsampling} together with the approach of \cite{giudice2023bayesian} to sample DAGs according to the GPN posterior and obtain the importance weights necessary for posterior inference on the network's features. In our simulations, we always assume all hyperparameters to be unknown and assign independent inverse-gamma priors to all lengthscales $\theta_j$ and noise standard deviations:
\begin{equation}
    \theta_j \,\sim\, \textrm{IG}(2, 2) \,,~\quad 
    \sigma \,\sim\, \textrm{IG}(1, 1). 
\end{equation}
We use the BiDAG \citep{bidag} hybrid implementation of the partition MCMC algorithm, and the RStan package \citep{rstan} for sampling from the hyperparameters' posterior distributions. Code to implement algorithms 1--3 and replicate all the simulations is available at \href{https://github.com/enricogiudice/causalGPNs}{\texttt{https://github.com/enricogiudice/causalGPNs}}.

\subsection{Results of MC Approach}
\label{sec:exps_glob}
\begin{figure}[!tbp]
\centering
\includegraphics[width = 0.75\textwidth]{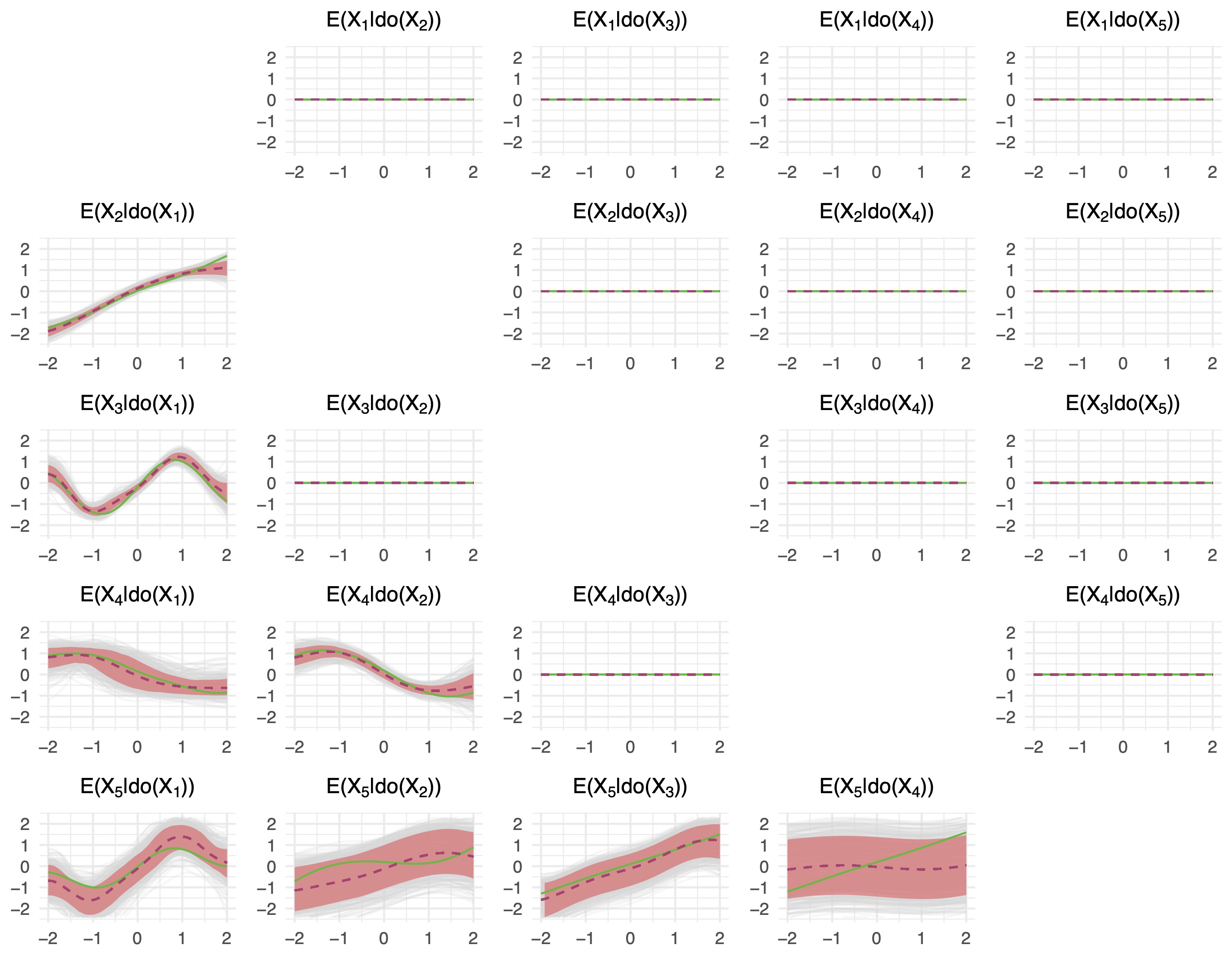} 
\caption{Estimated intervention expectations derived according to algorithm 1, with a known DAG. Samples are shown in gray, the dashed red line indicates the mean estimate, the green line shows the true data-generating value, and the red area shows an $80\%$ credible interval.}
\label{fig:glob_dag}
\end{figure}

In this subsection we showcase the results of the method to obtain intervention distributions outlined in section \ref{sec:glob}.
Figure \ref{fig:glob_dag} displays the result of algorithm 1 applied to data obtained with the generating process outlined in the previous subsection. 
The plot on the $\ell$-th row and $k$-th column of the figure shows the estimates of $\mathbb{E}(X_\ell\given \textrm{do}(X_k=x_k))$ as a function of different intervention levels $x_k$ on the x-axis, conditionally on knowledge of the DAG in figure \ref{fig:dag}. Each gray line represents a different sample from the posterior predictive distribution of $\mathbb{E}(X_\ell\given \textrm{do}(X_k=x_k))$ obtained with algorithm 1. The dotted red line for each plot shows the mean estimate, obtained by averaging the samples for every intervention level $x_k$. The red-shaded area covers $80\%$ of the posterior density. The continuous green line instead indicates the true value of $\mathbb{E}(X_\ell\given \textrm{do}(X_k=x_k))$ that was used to generate the data. Since we assume the DAG is known in this case, the panels of figure \ref{fig:glob_dag} show a line at zero whenever the intervention variable $X_k$ has no causal effect on $X_\ell$. This is because whenever there is no directed path from $X_k$ to $X_\ell$, algorithm 1 will estimate intervention distributions with the simple expectation $\mathbb{E}(X_\ell)$.

\begin{figure}[ht]
\centering
\includegraphics[width = 0.75\textwidth]{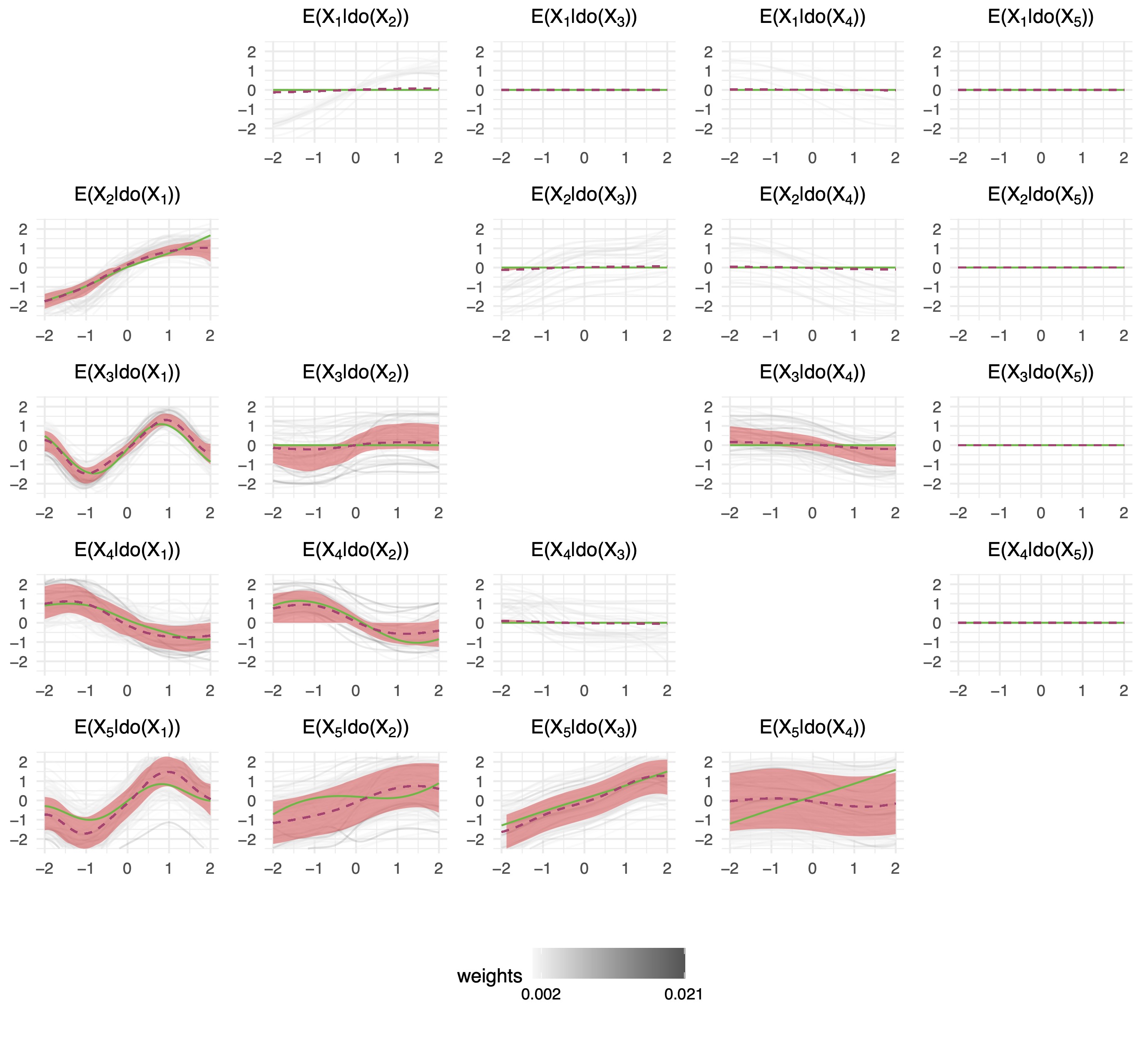} 
\caption{Estimated intervention expectations derived according to algorithm 2, without a known DAG. Samples are shown in gray, the dashed red line indicates the mean estimate, the green line shows the true data-generating value, and the red area shows an $80\%$ credible interval.}
\label{fig:glob_nodag}
\end{figure}

Figure \ref{fig:glob_nodag} shows the results without prior knowledge of the DAG, trained on the same data as figure \ref{fig:glob_dag}. In this case, we follow the procedure of sampling an ensemble $\mathcal{G}_1,...\,,\mathcal{G}_M$ of DAGs together with their associated weights $w_1,...\,,w_M$. Conditionally on each DAG $\mathcal{G}_m$, we generate a sample $x_{\ell,m}(x_k)$ from $\mathbb{E}(X_\ell\given \mathcal{G}_m, \textrm{do}(X_k=x_k))$. These samples are visible on the plots as lines, with their color varying from white to black depending on their associated  weight $w_m$. The dotted red mean line is computed as the weighted mean of the samples $x_{\ell,m}(x_k)$ for every intervention level $x_k$. 
The quantiles for the credible intervals are computed via the weighted Harrell–Davis estimator \citep{HarrellDavis, akinshin2023weighted}

Compared to figure \ref{fig:glob_dag}, the variance of the estimates increases considerably. This is due to the additional uncertainty regarding the DAG structure, which is added to the existing uncertainty regarding the parameters. The true intervention expectations however remain roughly within the $80\%$ credible interval.

\subsection{Local Approximation Results}
\begin{figure}[!tbp]
\centering
\includegraphics[width = 0.75\textwidth]{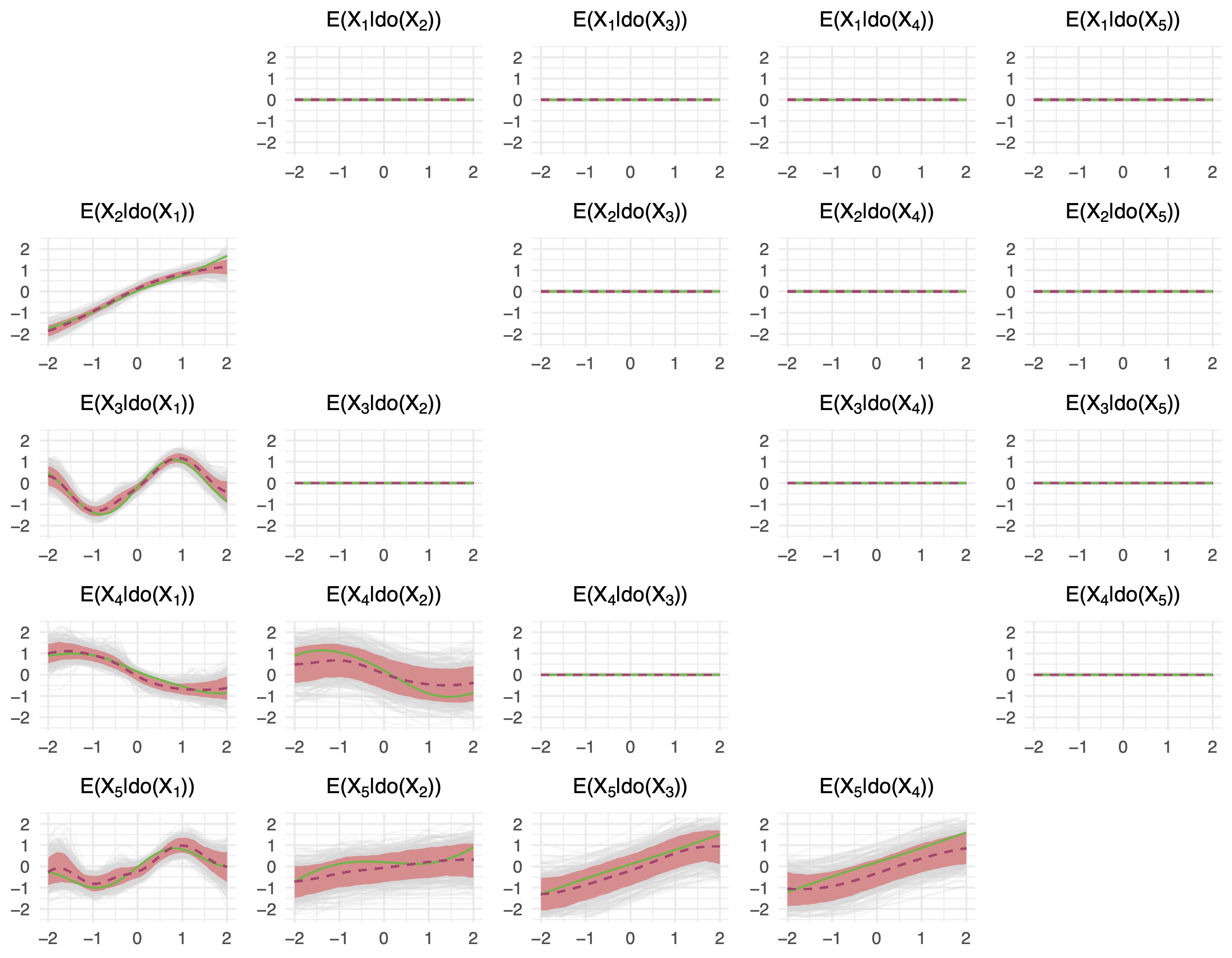} 
\caption{Estimated intervention expectations derived according to the local approximation with a known DAG. Samples are shown in gray, the dashed red line indicates the mean estimate, the green line shows the true data-generating value, and the red area shows an $80\%$ credible interval.}
\label{fig:loc_dag}
\end{figure}

We repeat the previous experiments with the local approximation outlined in section \ref{sec:loc}, which models the intervention distributions as a function of the outcome, intervention, and parent variables only. 
Figure \ref{fig:loc_dag} shows the results obtained from sampling from the posterior (\ref{marg_post}) assuming the DAG in figure \ref{fig:dag} is known, trained on the same data used for figures \ref{fig:glob_dag}--\ref{fig:glob_nodag}. The results are similar to those of the MC approach taken in figure \ref{fig:glob_dag}, although the local approximation tends to display more biased behaviours as the paths connecting the intervention and outcome variables increase in length.

\begin{figure}[!tbp]
\centering
\includegraphics[width = 0.75\textwidth]{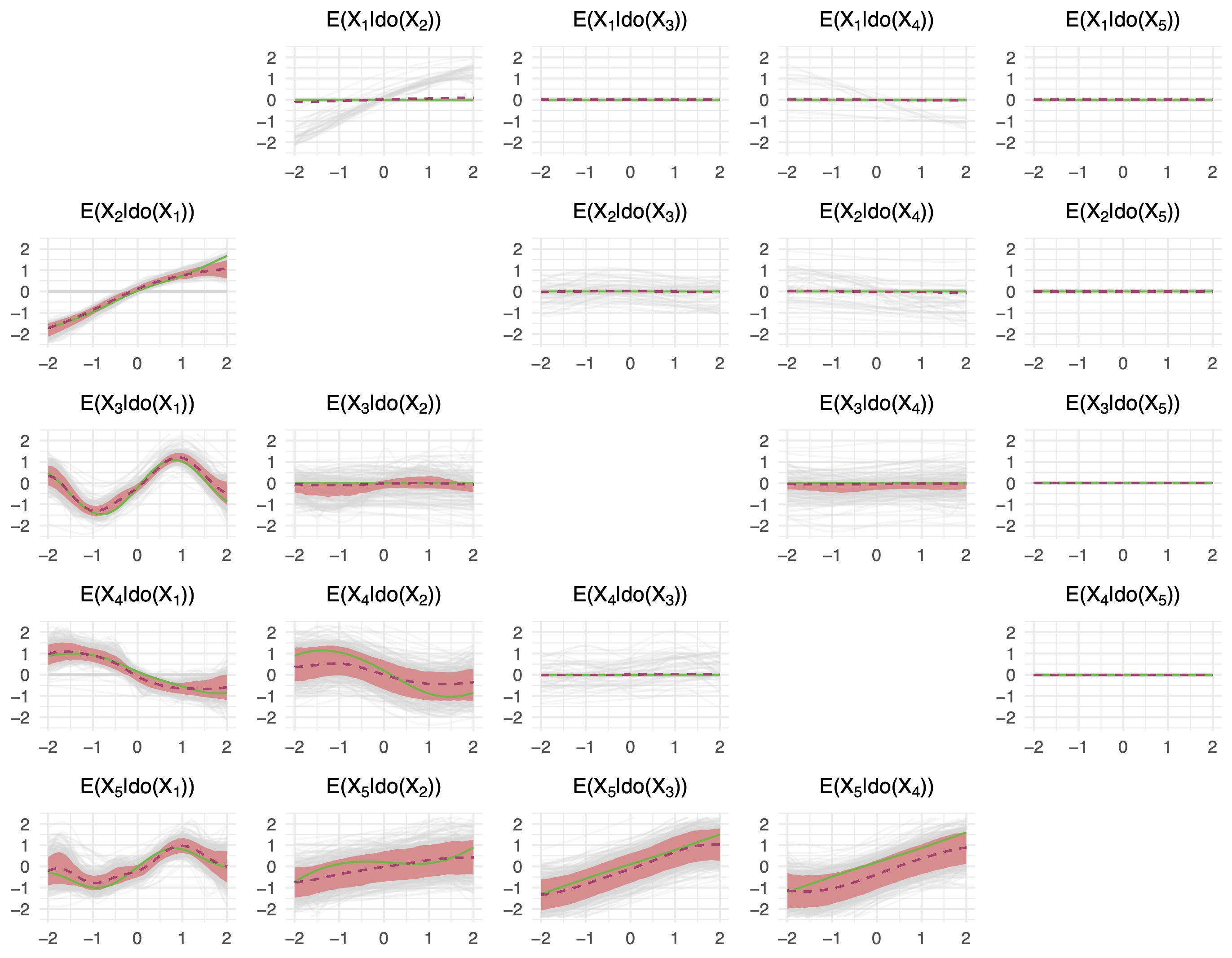} 
\caption{Estimated intervention expectations derived according to algorithm 3 without a known DAG. Samples are shown in gray, the dashed red line indicates the mean estimate, the green line shows the true data-generating value, and the red area shows an $80\%$ credible interval.}
\label{fig:loc_nodag}
\end{figure}

Figure \ref{fig:loc_nodag} shows the results of algorithm 3 trained on the same data without the known DAG assumption. As in the case seen in section \ref{sec:exps_glob}, the variance of the estimates increases without a noticeable increase in bias. The average number of sampled parents for every intervention-outcome variable pair is $2$, with minimum and maximum values of $1$ and $3$ respectively.

\begin{figure}[h]
  \centering
  \begin{subfigure}[c]{0.4\textwidth}
         \centering
    \includegraphics[width = \textwidth]{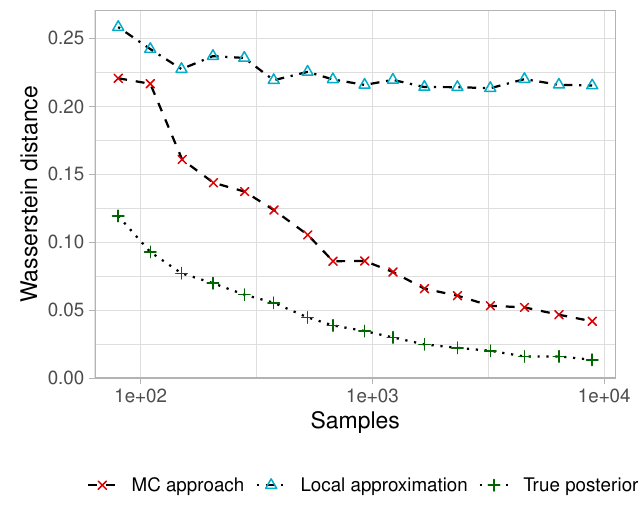}
  \end{subfigure}
    \hspace{0.2 cm}
  \begin{subfigure}[c]{0.4\textwidth}
    \centering
    \includegraphics[width = \textwidth]{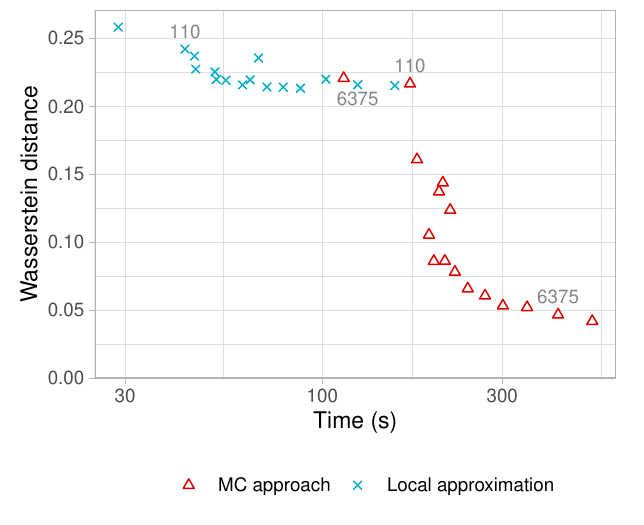}
  \end{subfigure}
  \centering
  \begin{subfigure}[c]{0.4\textwidth}
         \centering
    \includegraphics[width = \textwidth]{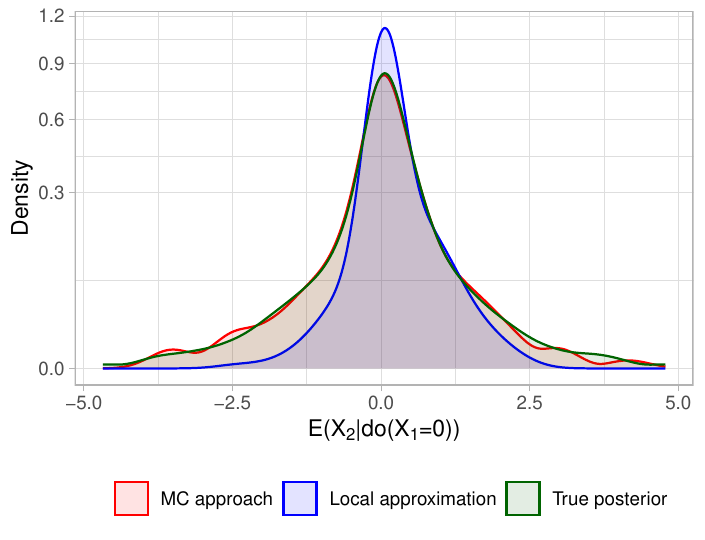}
  \end{subfigure}
  \hspace{0.2 cm}
  \begin{subfigure}[c]{0.4\textwidth}
    \centering
    \includegraphics[width = 0.6\textwidth]{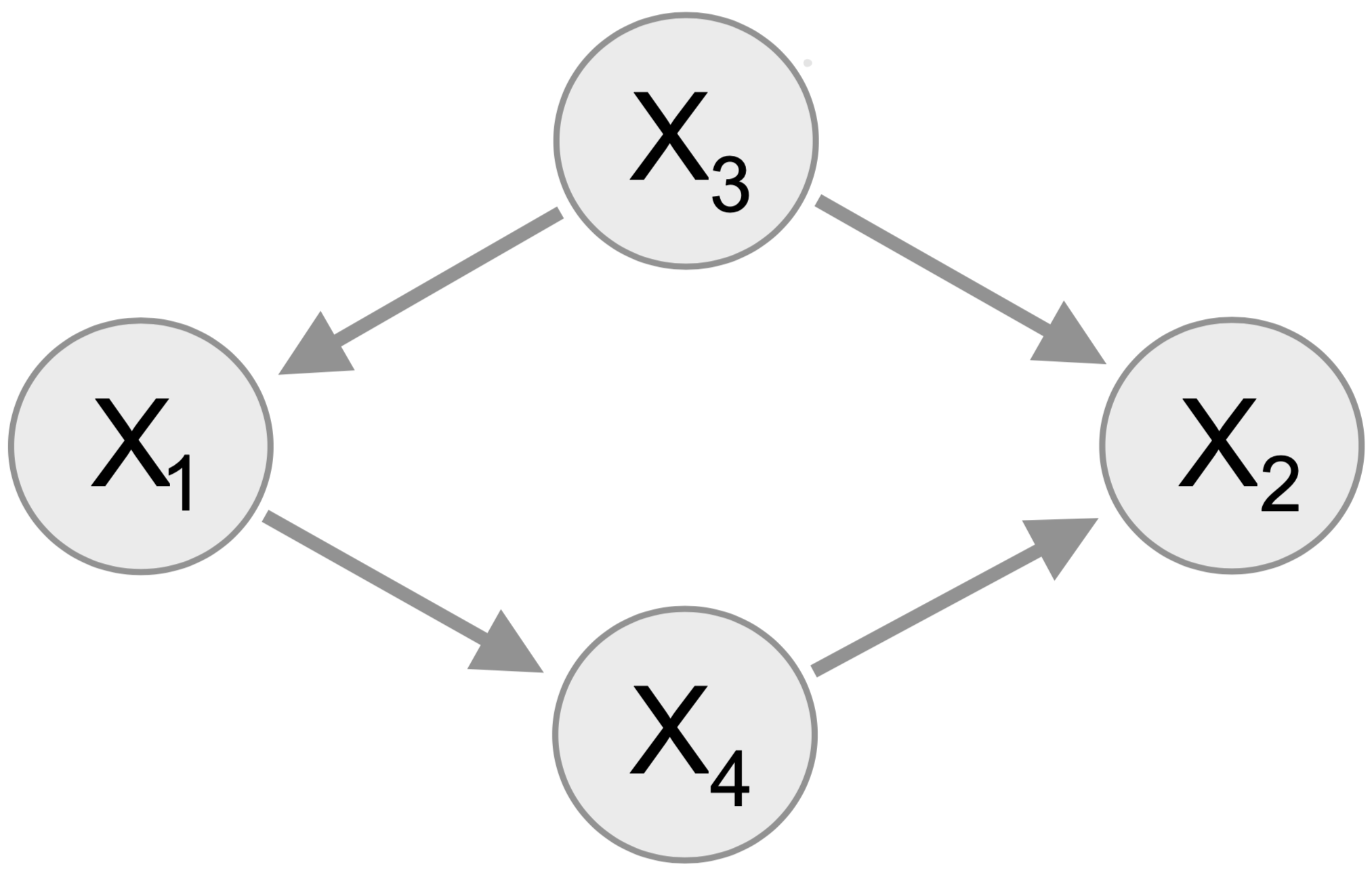}
  \end{subfigure}
  \caption{Top left: Wasserstein distance between the true posterior and the posterior obtained via the MC approach (red), and the posterior obtained via the local approximation (blue) as a function of the number of sampled DAGs. 
  Top right: the Wasserstein distances of the two approaches as a function of run-time. A selection of sample sizes are shown in gray.
  Bottom left: The estimated posteriors of the MC approach (in red) and local approximation (in blue) compared to the true posterior (in green). 
  Bottom right: the DAG used to generated the data.}
  \label{fig:post}
\end{figure}

In order to provide a direct comparison between the MC and local approaches, we examine the posterior distributions obtained by the two methods together with the ``true'' posterior. The true posterior is obtained by enumerating all possible DAGs and computing their posterior probability via bridgesampling \citep{bridgesampling}. A total of $10^4$ samples of the intervention expectation of interest are then collected from the DAGs proportionally to their posterior probability. Since enumerating all possible DAGs is only feasible for small graphs, we consider a DAG with $4$ nodes. We generate $100$ data samples according to equation (\ref{Fourier}) from the DAG in the bottom right panel of figure \ref{fig:post}. We then compute the posteriors of the intervention expectation $\mathbb{E}(X_2\given \textrm{do}(X_1=0))$ using the MC and local approaches, always assuming the underlying DAG to be unknown.

The top left panel of figure \ref{fig:post} shows the first-order Wasserstein distance between the estimated and true posterior samples of $\mathbb{E}(X_2\given \textrm{do}(X_1=0))$. For a given set of observations $\{x_1,...\,,x_M\}$ and corresponding weights $\{w_1,...\,,w_M\}$, we compute the (weighted) empirical distribution function $\hat{F}(t) = \sum_{i=1}^M w_i \scalebox{1.1}{$\boldone$}\scalebox{0.8}{$(x_i\le t)$}$. We then calculate the Wasserstein distance between the estimated posterior distribution function $\hat{F}(t)$ and the true posterior distribution function $G(t)$ as
\begin{equation}
    W(\hat{F}, G) \,=\, \int \vert \hat{F}(t) - G(t) \vert \,dt.
\end{equation}
For both the MC and local approaches, the Wasserstein distances are shown as a function of the number of samples $M$, averaged over $50$ runs on the same data. Since the true posterior is itself estimated with samples, we show the average distance between different realizations of the posterior, which provides a lower bound for the Wasserstein distance between our methods and the true posterior. The results indicate that the distance between the posterior distribution estimated by the MC approach and the true posterior decreases as the number of samples is increased and the estimates become more accurate. The distance of the estimate derived from the local approximation is however higher, indicating lack of convergence of the method to the true posterior distribution. This supports the intuition that the additive GP model introduces some degree of model misspecification and is unable to provide a highly accurate representation of the true posterior intervention distribution.

The top right panel of figure \ref{fig:post} shows the same Wasserstein distances of the top left panel as a function of the average run-time of the different algorithms. The local approximation is significantly more efficient, running on average around $3.5$ times faster for every level of $M$ than the MC approach. 

The bottom left panel shows the kernel density estimates based on $10^4$ samples from the posteriors for each of the two approaches. The MC (in red) and local (in blue) approaches are compared to the true posterior in green. The results indicate that both methods are able to adequately capture the bulk of the posterior, although the local approximation underestimates the variance of the target distribution.

\section{Application to Gene Regulatory Networks}
\label{sec:apps}
In this section we apply our GP-based intervention estimation method to analyze the effects of hypothetical interventions on the gene expression levels in \textit{Arabidopsis thaliana} \citep{wille2004}. The publicly available dataset consists of $118$ different observations of $39$ genes obtained via microarrays. In our analysis, we focus on the MEV pathway described by the $n=13$ genes that were previously studied by \cite{Castelletti2021}. After log-transforming and standardizing the data, the marginal distributions of the variables exhibit various degrees of non-Gaussian behavior, indicating that adopting a non-parametric approach is appropriate. 

\begin{figure}[t]
\centering
\includegraphics[width = 0.75\textwidth]{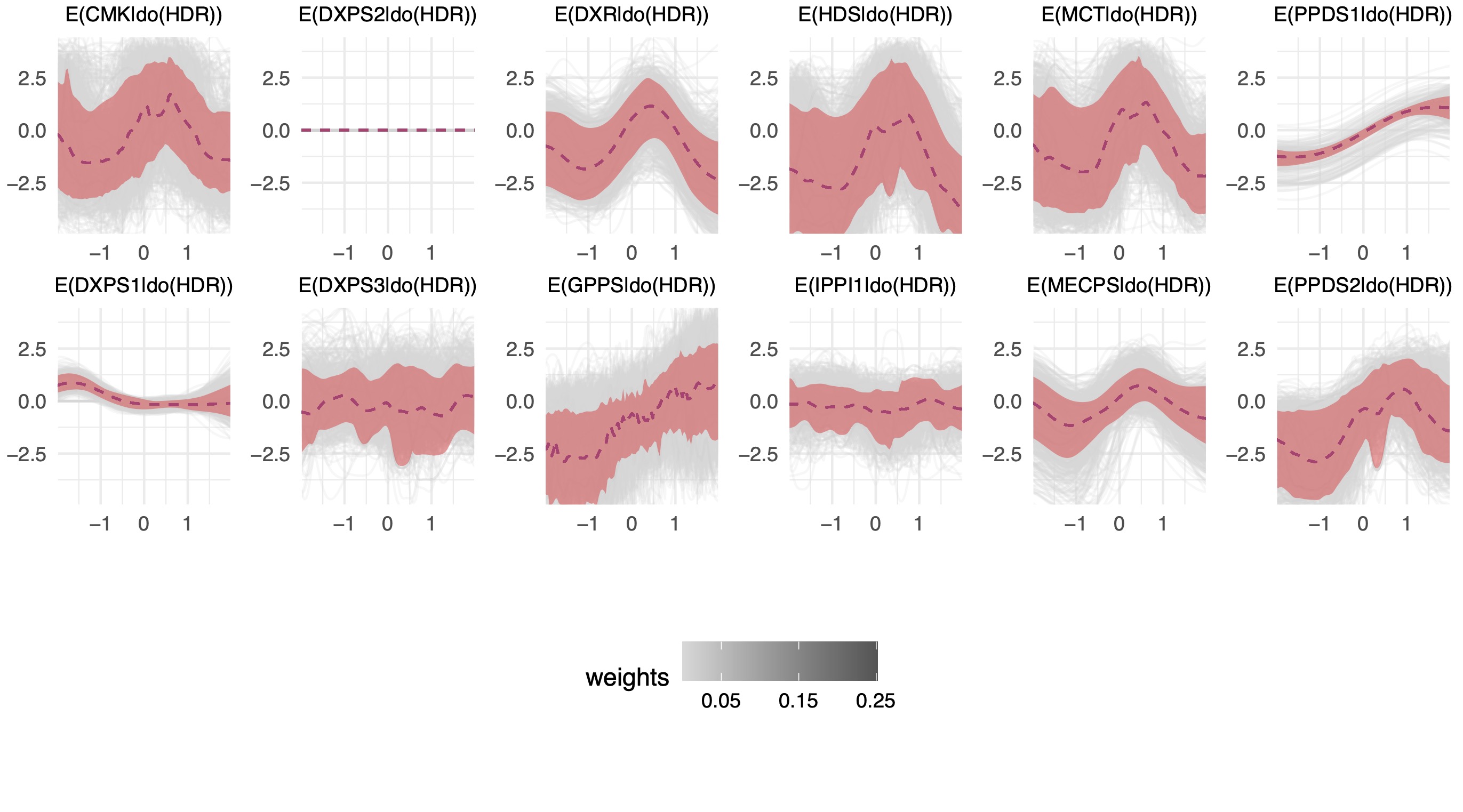} 
 
\includegraphics[width = 0.75\textwidth]{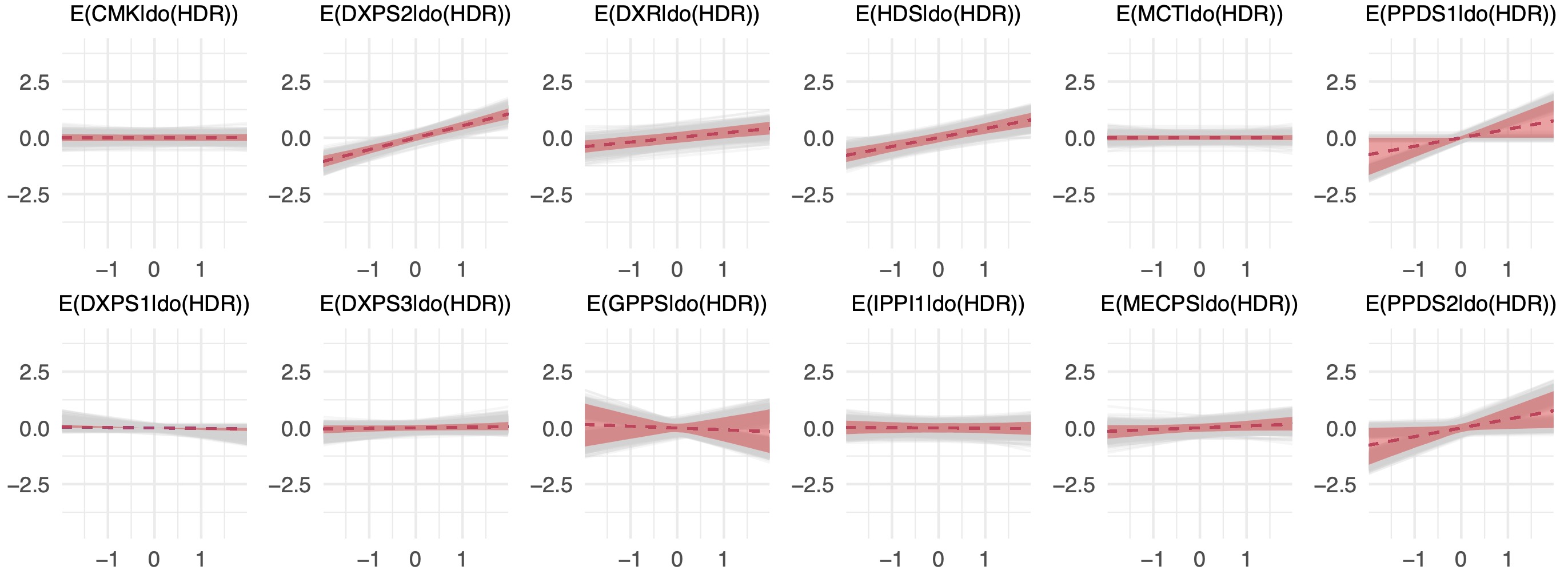} 
\caption{Results of an intervention on the variable \textit{HDR} from the GPN MC-based approach (above) and from the Gaussian-linear approach (below). Samples are shown in gray, the dashed red line indicates the mean estimate, and the red area shows an 80\% credible interval.}
\label{fig:hdr}
\end{figure}

We follow the MC approach described in section \ref{sec:nodag_glob}, sampling DAGs from the GPN posterior and simulating the effect of different interventions across the sampled network. We sample $M=800$ DAGs from their posterior distribution via the partition MCMC algorithm \citep{partitionmcmc} and visualize the estimated expectation of the intervention distributions following the example of figure \ref{fig:glob_nodag}. For each variable we consider a range of intervention values and plot the (weighted) sampled functions together with the mean estimate and an $80\%$ credible interval. The top two rows of plots in figure \ref{fig:hdr} show the estimated posterior distributions on all variables for interventions on the \textit{HDR} variable. In this example, different intervention levels of the gene are predicted to have a substantial impact on the distributions of most of the other variables. The results identify a diverse collection of linear and non-linear causal relations between the intervention and outcome variables. 

The bottom two rows of plots in figure \ref{fig:hdr} allow comparing the results of the GPN model with the results of a linear model such as the approach of \cite{Castelletti2021}, where the variables are assumed to be jointly Gaussian. In such a model, intervention distributions among genes can be estimated following a Bayesian approach; DAGs are sampled according to the BGe score \citep{Geiger2002,kuipers2014addendum} and the parameters of the joint distribution are sampled according to their posterior distribution \citep{viinikka2020}. The Gaussian assumption implies that the sampled functions from the posterior of $\mathbb{E}(Y \given \textrm{do}(X=x))$ are linear in $x$.

\begin{figure}[t]
\centering
\includegraphics[width = 0.75\textwidth]{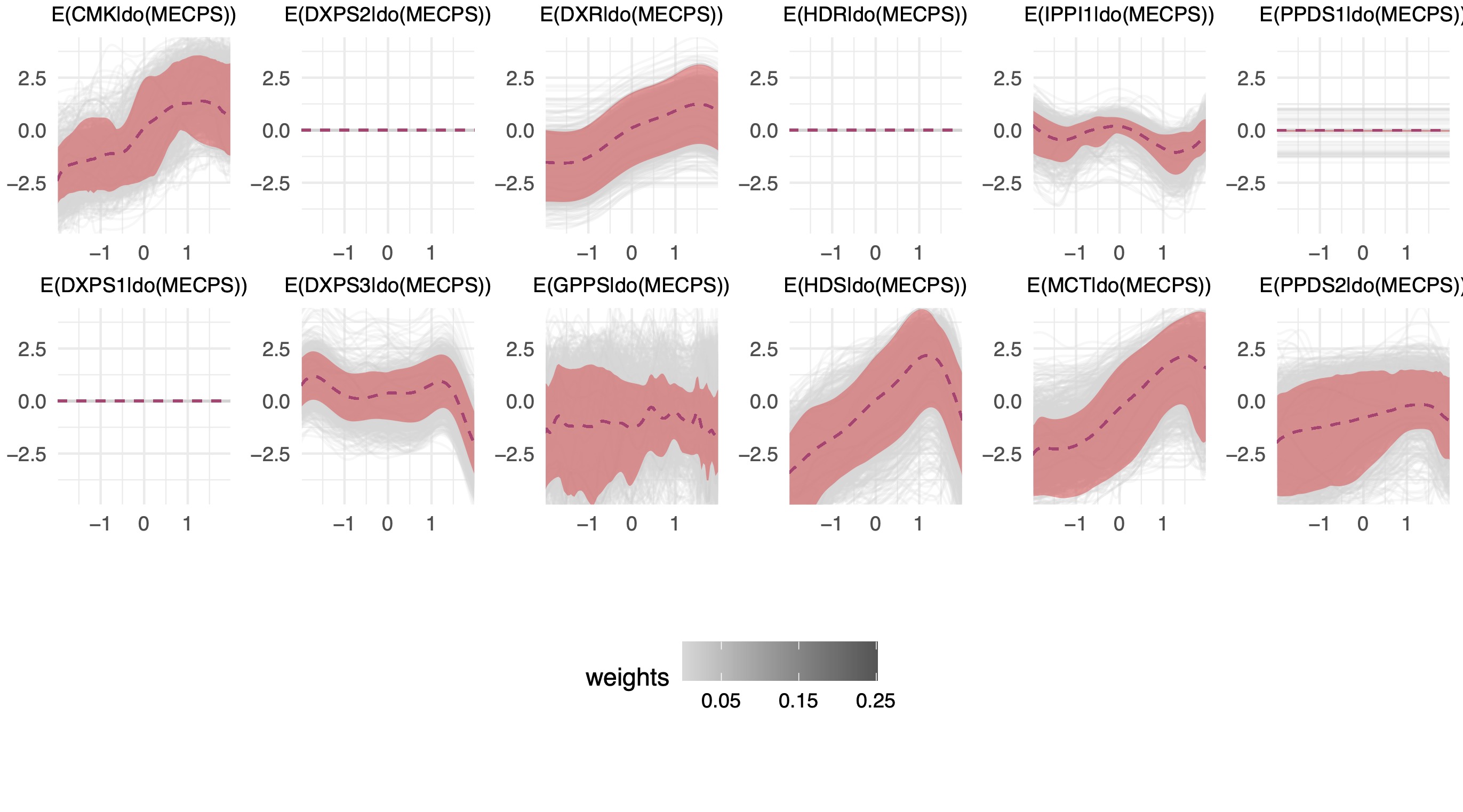} 
 
\includegraphics[width = 0.75\textwidth]{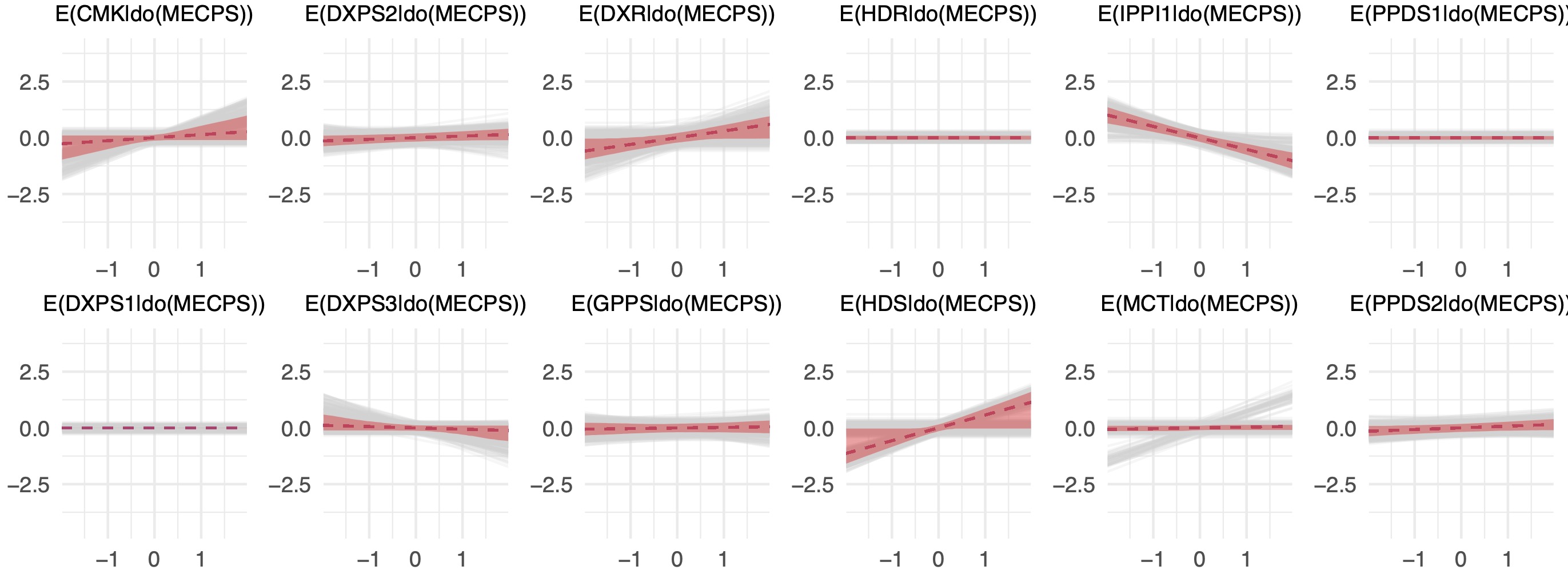} 
\caption{Results of an intervention on the variable \textit{MECPS} from the GPN MC-based approach (above) and from the Gaussian-linear approach (below). Samples are shown in gray, the dashed red line indicates the mean estimate, and the red area shows an 80\% credible interval.}
\label{fig:mecps}
\end{figure}

The plots in figure \ref{fig:mecps} show the results of the GPN (above) and linear-Gaussian (below) models when intervening on the \textit{MECPS} gene. Also in this case, the GPN analysis reveals a rich variety of non-linear causal effects which depend on the different intervention levels. The results from the linear model are in general agreement with those obtained with our GPN model but lack the ability to identify the variety of non-linear relationships. The linear-Gaussian model also outputs a lower uncertainty in the posteriors, reflecting its stricter assumptions compared to the nonparametric GPN model. This behavior can also be seen in figure \ref{fig:bestie_nodag} in the supplementary material, which shows the linear-Gaussian results for the same simulation setting of figure \ref{fig:glob_nodag}. Non-linear relations are typically not identified by the linear model, which will estimate a narrow confidence interval around its biased MAP estimate.

The high variance of many posterior estimates reflect a situation of high uncertainty regarding the causal structure of the pathway. Nevertheless, the posteriors of specific intervention/outcome gene pairs are highly appealing since our model predicts vastly different results than the linear model and these genes may therefore represent an interesting target for intervention experiments.

\begin{figure}[!tbh]
\centering
\includegraphics[width = 0.496\textwidth]{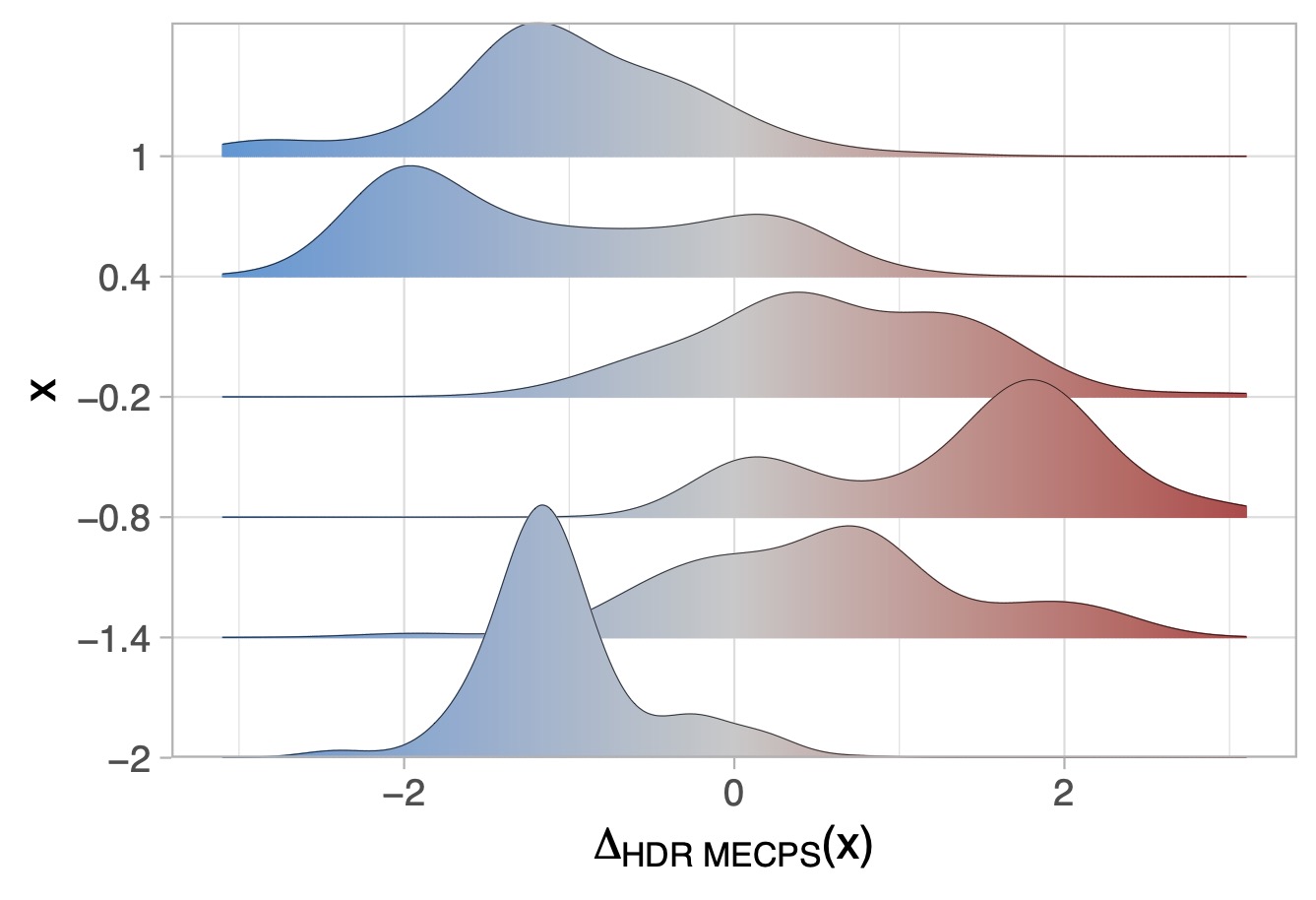} 
\includegraphics[width = 0.496\textwidth]{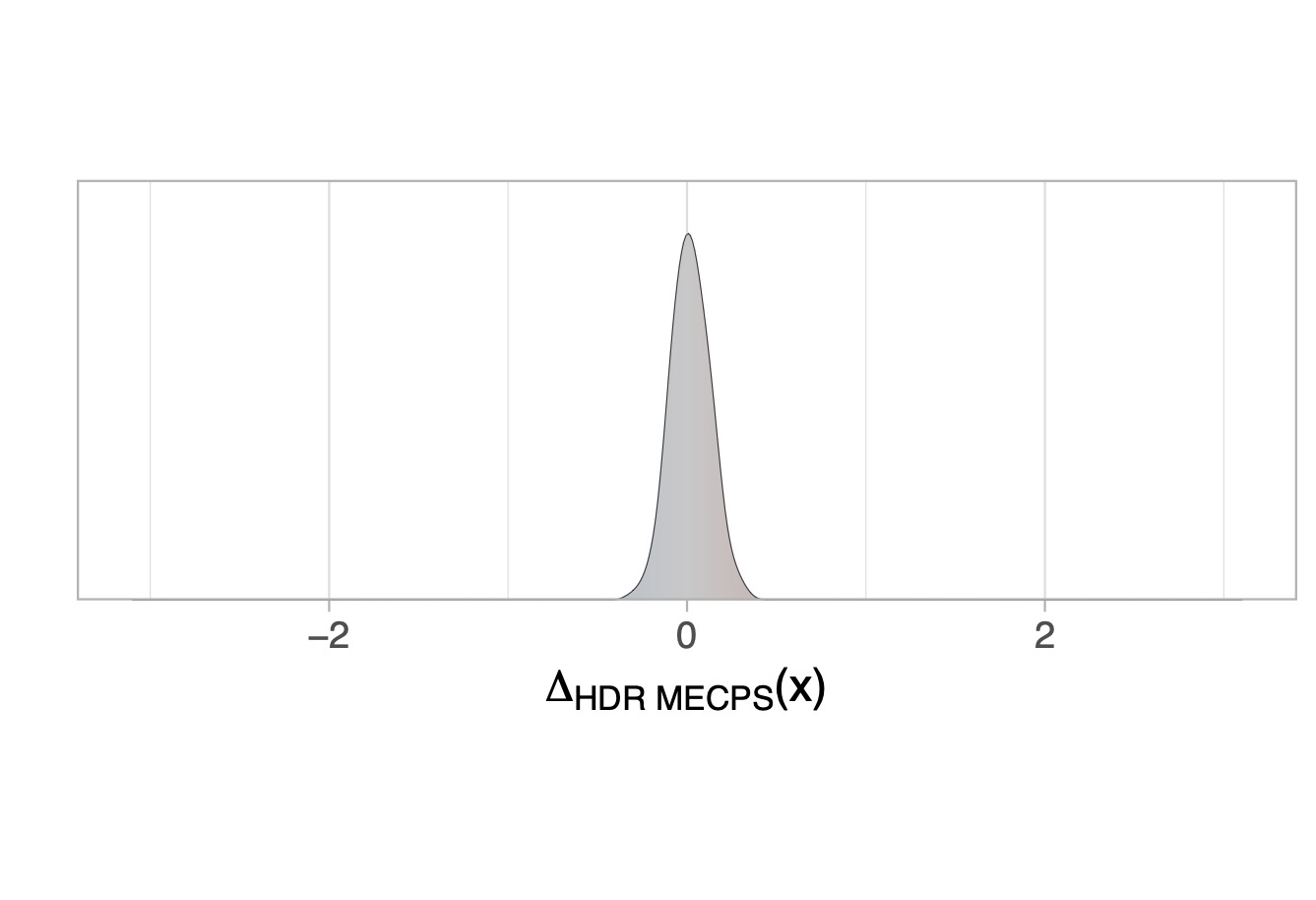} 
\caption{Density plots showing the estimated distributions of the causal effect of \textit{HDR} on \textit{MECPS} for a selection of intervention values $x$. On the left are the estimates obtained with the GPN model, on the right the results of the linear model. Red and blue values show respectively positive and negative estimates.}
\label{fig:joy}
\end{figure}

We can additionally analyze causal effects, i.e.~the change in the target expectation of the intervention distribution when comparing two different intervention levels. For an outcome variable $Y$ and an intervention variable $X$, we are interested in the difference 
\begin{equation}
    \Delta_{XY}(x) \,\coloneqq\, \mathbb{E}(Y \given \textrm{do}(X=x+1)) - \mathbb{E}(Y \given \textrm{do}(X=x))\,.
\end{equation}
The left panel of figure \ref{fig:joy} shows the estimated posterior densities of $\Delta_{XY}(x)$ for the outcome variable \textit{MECPS} and the intervention variable \textit{HDR}, for different values of the intervention level $x$. On the right panel is the posterior distribution for the (constant) causal effect in the linear-Gaussian model, which corresponds to the posterior of the slope coefficient of the sampled functions in the corresponding panel of figure \ref{fig:hdr}. The GPN model identifies a variety of casual effects, ranging from positive to zero depending on the intervention value $x$. The results of the linear model are in agreement with the GPN results, but the different causal effects are averaged together into one parameter whose posterior distribution is peaked around a single value.

The results of the interventions on the variables \textit{HDR} and \textit{MECPS} from the local approximation are deferred to the supplementary material \ref{supp}. 

\section{Conclusions}
In this work we have studied the problem of estimating intervention distributions from observational data in GPNs, both with and in absence of knowledge of the underlying DAG. Causal inference can be approached as a global process, taking into account all upstream variables in the graph, or as a function of local variables only. We have shown the way in which either of these approaches can be taken to perform causal inference with the GPN model. While the local approximation can be potentially misspecified due to the modeling assumptions in equation (\ref{full_addmodel}), it allows more efficient causal inference that is suitable for large networks in which propagating samples can be computationally expensive. Experimental results indicate that both methods are able to provide a reasonably similar posterior over the intervention distributions. An interesting future direction of research would be to extend the additive GP framework (\ref{full_addmodel}) to a deep GP model \citep{deepGPs}, which might better express the complex relationships between intervention and outcome variables in GPNs.

For descriptive purposes we have employed a number of simplifying modeling choices, such as the additive squared exponential kernel function (\ref{addkernel}) or Gaussian marginal distributions for root nodes. These choices allow for interesting extensions to more general models, such as higher-order additive kernels \citep{additiveGPs} or modeling root nodes with mixtures of Gaussians \citep{Mooij2010, GPNinterventions}. The additive model (\ref{part_addmodel}) employed in the local approximation has the benefit of being easily interpretable, but can prove inadequate for large networks with complex dependencies, requiring more general models, such as higher-order additive kernels \citep{additiveGPs}.

In this work we have focused exclusively on interventions performed on single variables. The truncated Markov factorization can however be applied for joint interventions involving several variables \citep{Pearlmut}. In this case the propagation of the samples (\ref{propagate}) down the interventional GPN remains unchanged when $X$ is a vector of variables being intervened upon. Extending the procedure to the case with an unknown DAG remains unchanged. 

The local approximation based on the backdoor adjustment formula however fails when considering joint interventions, since a valid backdoor set may not exist under such circumstances \citep{Pearl2000}. Because of this, the MC approach is the most viable option when considering interventions on multiple variables. An alternative method would involve computing the effects of interventions on sets of variables with only knowledge of the parent set of that intervention set, such as the recursive regression approach proposed by \cite{Nandy2017}. Extending such an approach to GPNs represents an interesting avenue for future research, and would offer valuable advancements on the analysis of efficient causal inference in high-dimensional GPNs.

\clearpage
\bibliographystyle{abbrvnat} 
\bibliography{references}

\clearpage
\appendix
\counterwithin{figure}{section}
\pagenumbering{arabic}

\section{Supplementary Material}
\label{supp}
\begin{figure}[h]
\centering
\includegraphics[width = 0.8\textwidth]{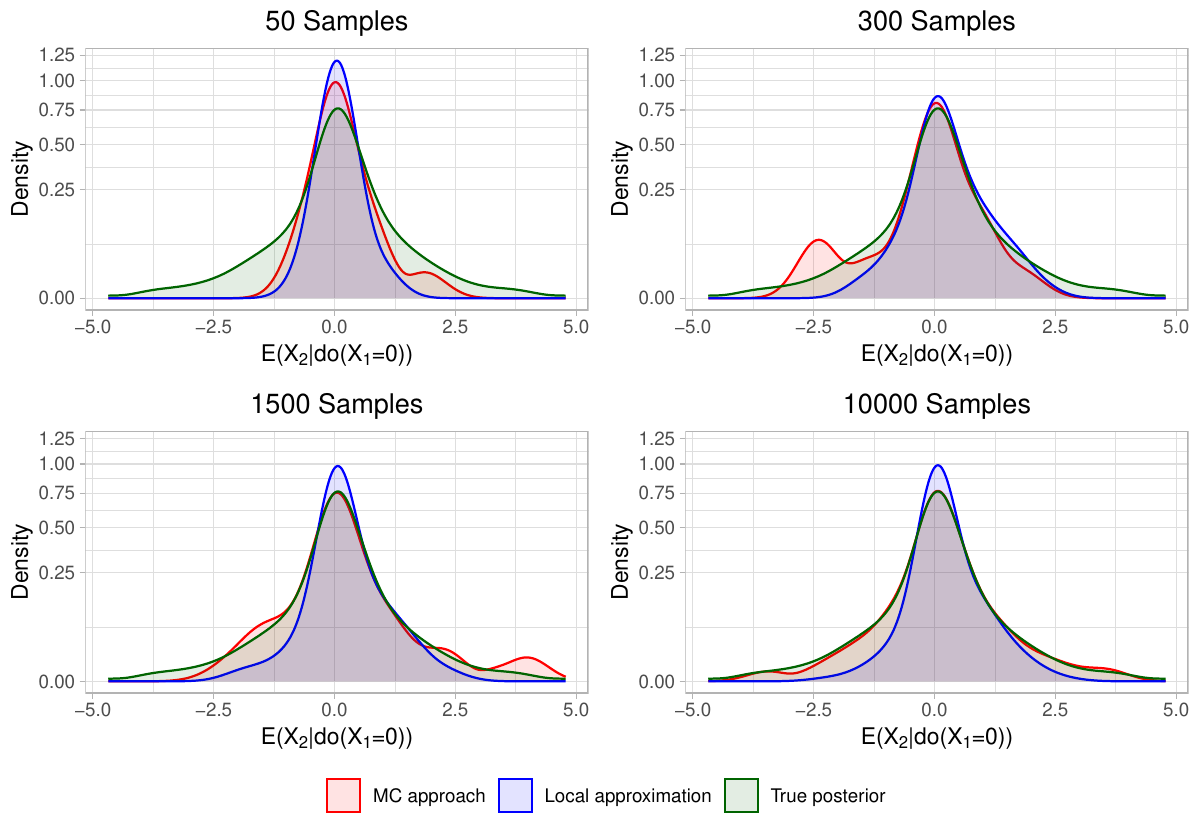} 
\caption{The estimated posteriors of the MC approach (in red) and local approximation (in blue) compared to the true posterior (in green) for different numbers of sampled DAGs in the experiment of figure \ref{fig:post}.}
\label{fig:denss}
\end{figure}

Here we provide additional results from the simulated and real datasets. Figure \ref{fig:denss} shows the kernel density estimates corresponding to different sample sizes in figure \ref{fig:post}. As the samples from the posterior increase, the results of the MC approach converge to the posterior of interest. The local approximation on the other hand under-estimates the variance of the intervention distribution, which is only in part improved by increasing the samples from the posterior.

\begin{figure}[h]
\centering
\includegraphics[width = 0.8\textwidth]{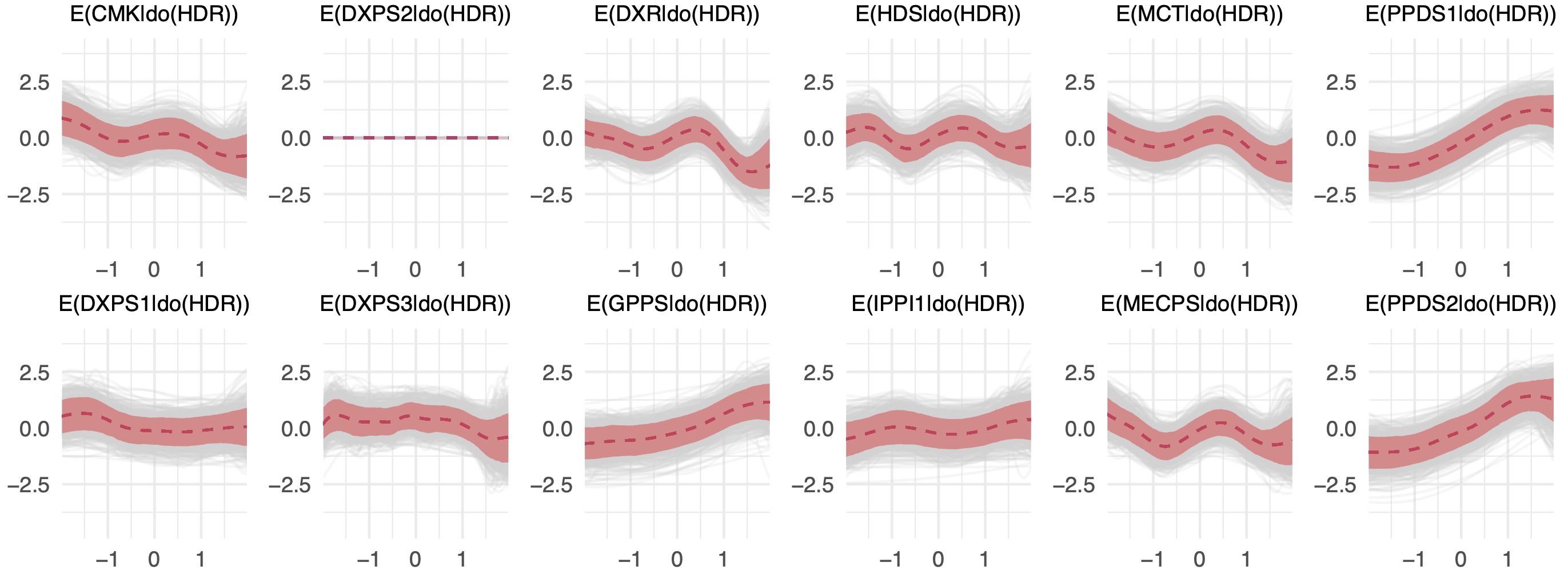} 
\caption{Results of an intervention on the variable \textit{HDR} from the GPN local approximation. Samples are shown in gray, the dashed red line indicates the mean estimate, and the red area shows an 80\% credible interval.}
\label{fig:HDRloc}
\end{figure}

\begin{figure}[h]
\centering
\includegraphics[width = 0.8\textwidth]{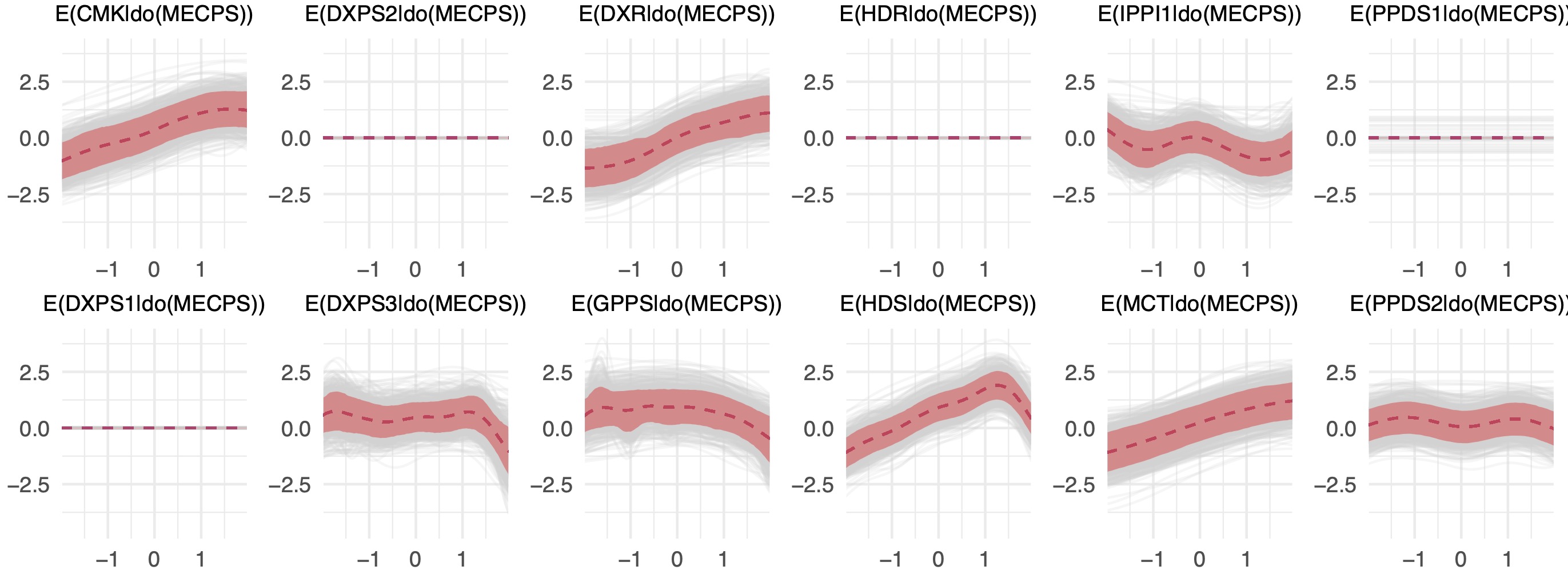} 
\caption{Results of an intervention on the variable MECPS from the GPN local approach. Samples are shown in gray, the dashed red line indicates the mean estimate, and the red area shows an 80\% credible interval.}
\label{fig:MECPSloc}
\end{figure}

Figures \ref{fig:HDRloc} and \ref{fig:MECPSloc} show the results of the local approximation trained on the \textit{A.\ thaliana} dataset for interventions respectively on the \textit{HDR} and \textit{MECPS} genes.

\begin{figure}[h]
\centering
\includegraphics[width = 0.75\textwidth]{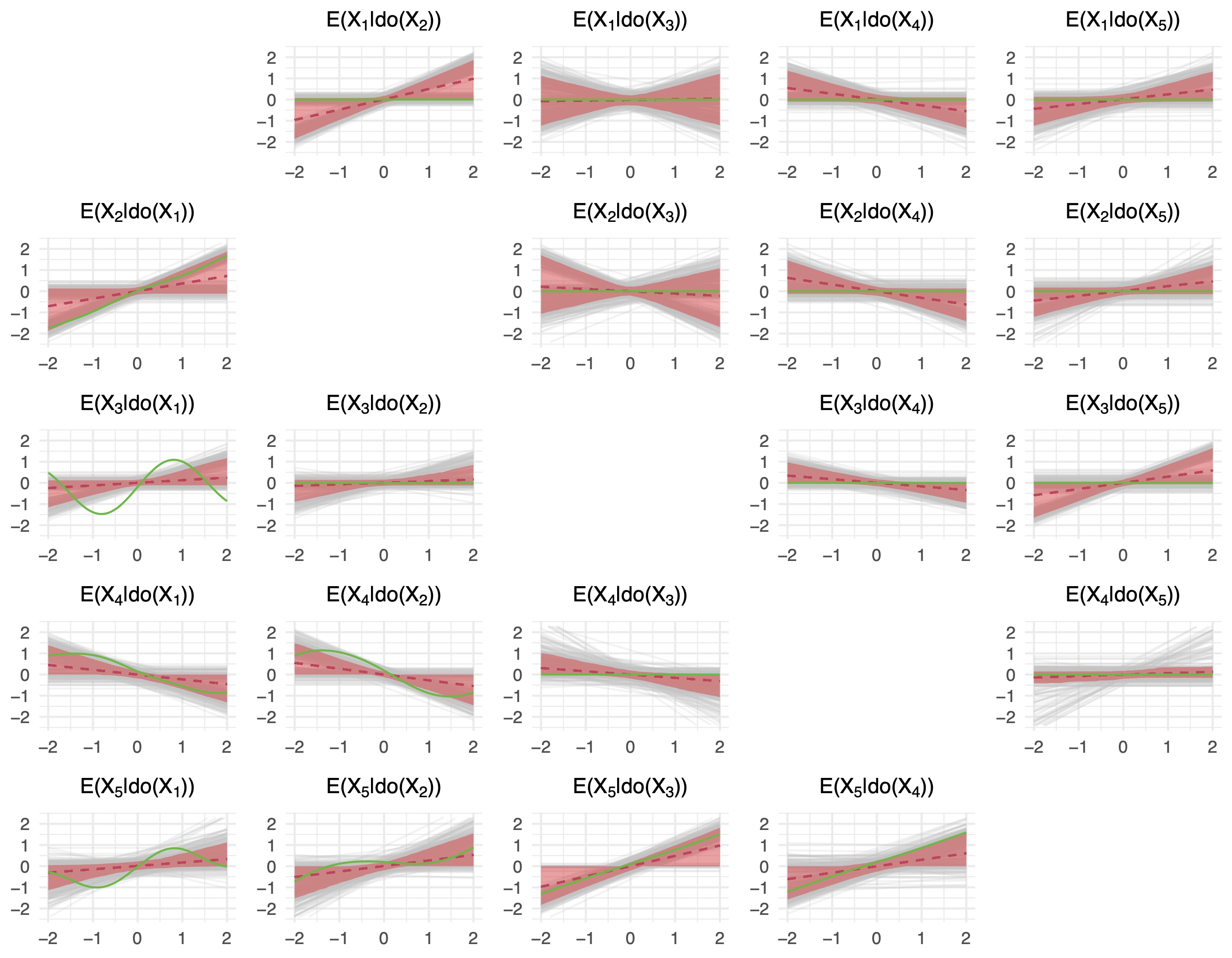} 
\caption{Estimated intervention expectations derived with the linear-Gaussian approach, without a known DAG. Samples are shown in gray, the dashed red line indicates the mean estimate, the green line shows the true data-generating value, and the red area shows an $80\%$ credible interval.}
\label{fig:bestie_nodag}
\end{figure}

Figure \ref{fig:bestie_nodag} shows the results of the linear-Gaussian approach trained on the same data as figure \ref{fig:glob_nodag}, without prior knowledge of the true DAG. The linear model is appropriate when the causal effects are linear, but results in strongly biased estimates when the true underlying causal relation (in green) is non-linear. In such cases, the model also underestimates the variability in the intervention distribution, which is reflected by a narrow credible interval.

\end{document}